\documentclass[10pt,twocolumn,letterpaper]{article}

\usepackage{cvpr}              
\usepackage[accsupp]{axessibility} 

\usepackage{graphicx}
\usepackage[table, svgnames, dvipsnames]{xcolor}

\usepackage{amsmath,amsthm,amsfonts,bm,physics,mathtools}
\usepackage{cancel}
\usepackage{algorithm}

\usepackage{algpseudocode}

\def\1{\bm{1}}

\def\eps{\bm{\epsilon}}

\def\rvx{{\mathbf{x}}}

\def\rvz{{\mathbf{z}}}

\def\vzero{{\bm{0}}}

\def\mI{{\bm{I}}}

\def\mO{{\bm{O}}}

\DeclareMathAlphabet{\mathsfit}{\encodingdefault}{\sfdefault}{m}{sl}
\SetMathAlphabet{\mathsfit}{bold}{\encodingdefault}{\sfdefault}{bx}{n}

\def\gL{{\mathcal{L}}}

\def\gN{{\mathcal{N}}}

\newcommand{\E}{\mathbb{E}}

\newcommand{\R}{\mathbb{R}}

\newcommand{\parderiv}[2]{\frac{\partial}{\partial #2} #1}
\newcommand{\gradnd}[2]{\nabla_{#2} #1}

\usepackage{array}
\usepackage{multirow}
\usepackage{booktabs}
\usepackage{color, colortbl}
\usepackage{makecell}
\usepackage{float}
\usepackage{thm-restate}
\usepackage{diagbox}
\usepackage{tikz}
\usepackage{fancyvrb}

\theoremstyle{plain}

\theoremstyle{definition}

\theoremstyle{remark}

\definecolor{Gray}{gray}{0.85}
\definecolor{LightCyan}{rgb}{0.88,1,1}

\makeatletter
\usepackage{xspace}
\def\@onedot{\ifx\@let@token.\else.\null\fi\xspace}
\DeclareRobustCommand\onedot{\futurelet\@let@token\@onedot}

\newcommand{\figref}[1]{Figure~\ref{#1}}
\newcommand{\equref}[1]{Eq\onedot~\eqref{#1}}

\newcommand{\tabref}[1]{Table~\ref{#1}}

\newcommand{\appref}[1]{Appendix~\ref{#1}}

\newcommand{\algoref}[1]{Algorithm~\ref{#1}}

\def\eg{\emph{e.g}\onedot}

\def\ie{\emph{i.e}\onedot}

\def\wrt{w.r.t\onedot}

\newcommand{\mypara}[1]{\noindent\textbf{#1}}

\newcommand{\model}[1][]{DreamPropeller}

\usepackage{multicol}

\usepackage{cuted}
\usepackage{capt-of}
\definecolor{cvprblue}{rgb}{0.21,0.49,0.74}
\usepackage[pagebackref,breaklinks,colorlinks,citecolor=cvprblue]{hyperref}

\def\paperID{11342} 
\def\confName{CVPR}
\def\confYear{2024}

\title{DreamPropeller: Supercharge Text-to-3D Generation with Parallel Sampling}

\author{Linqi Zhou$^1$\thanks{Work done at Pika Labs.} \quad\quad\quad\quad Andy Shih$^1$ \quad\quad\quad\quad Chenlin Meng$^{1,2}$ \quad\quad\quad\quad Stefano Ermon$^1$\\
$^1$Stanford University, $^2$Pika Labs\\
{\tt\small $^1$\{linqizhou, andyshih, chenlin, ermon\}@stanford.edu, $^2$chenlin@pika.art}
}

\begin{document}
\maketitle


\def\arxiv{1}

\everypar{\looseness=-1}

\begin{abstract}

Recent methods such as  Score Distillation Sampling (SDS) and Variational Score Distillation (VSD) using 2D diffusion models for text-to-3D generation
have demonstrated impressive generation quality. However, the long generation time of such algorithms significantly degrades the user experience.
To tackle this problem, we propose \model, a drop-in acceleration algorithm that can be wrapped around any existing text-to-3D generation pipeline based on score distillation. 
Our framework generalizes \textit{Picard iterations}, a classical algorithm for parallel sampling an ODE path, and can account for non-ODE paths such as momentum-based gradient updates and changes in dimensions during the optimization process as in many cases of 3D generation.  We show that our algorithm trades parallel compute for wallclock time and empirically achieves up to 4.7x speedup with a negligible drop in generation quality for all tested frameworks.
Our implementation can be found \href{https://github.com/alexzhou907/DreamPropeller}{here}.
\end{abstract}

\begin{figure}
    \centering
    \setlength{\tabcolsep}{0pt}
    \begin{tabular}{@{}>{\kern-\tabcolsep}c@{}c@{}c<{\kern-\tabcolsep}@{}}
  & DreamGaussian \citep{Tang_undated-od} & ProlificDreamer \citep{Wang2023-lp} \\
     \parbox[t]{4mm}{\rotatebox[origin=c]{90}{(Complete)}} & 
    \begin{minipage}{.5\linewidth}
          \includegraphics[width=\linewidth,trim=1 1 1 1,clip]{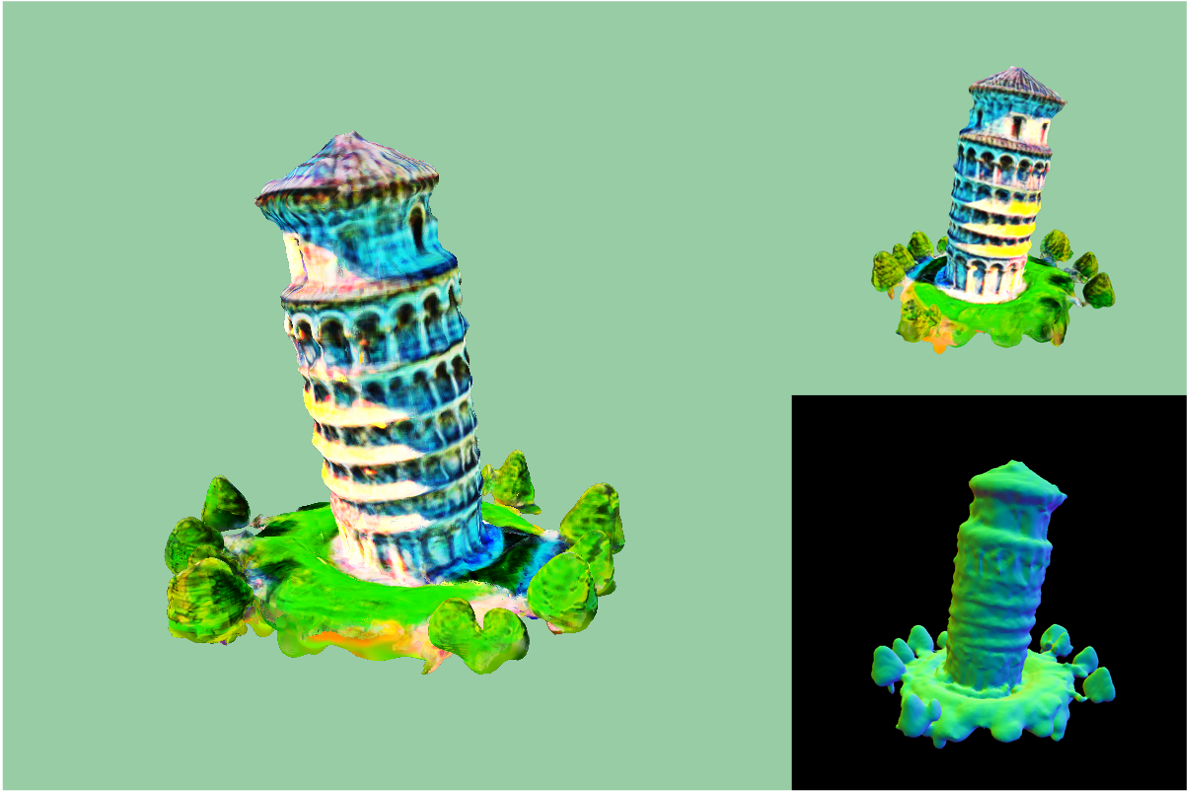}
        \end{minipage}
        &
         \begin{minipage}{.5\linewidth}
          \includegraphics[width=\linewidth,trim=1.5 1.5 1.5 1.5,clip]{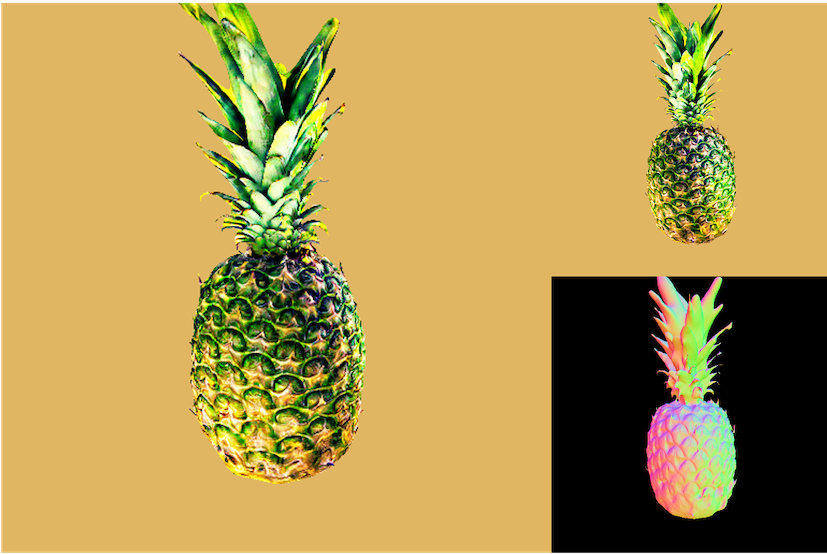}
        \end{minipage} \\
      & \cellcolor{Gainsboro!60}11 minutes &\cellcolor{Gainsboro!60} 9 hours 28 minutes \\
  \parbox[t]{4mm}{\rotatebox[origin=c]{90}{(Incomplete)}}  & \begin{minipage}{.5\linewidth}
          \includegraphics[width=\linewidth,trim=1 1 1 1,clip]{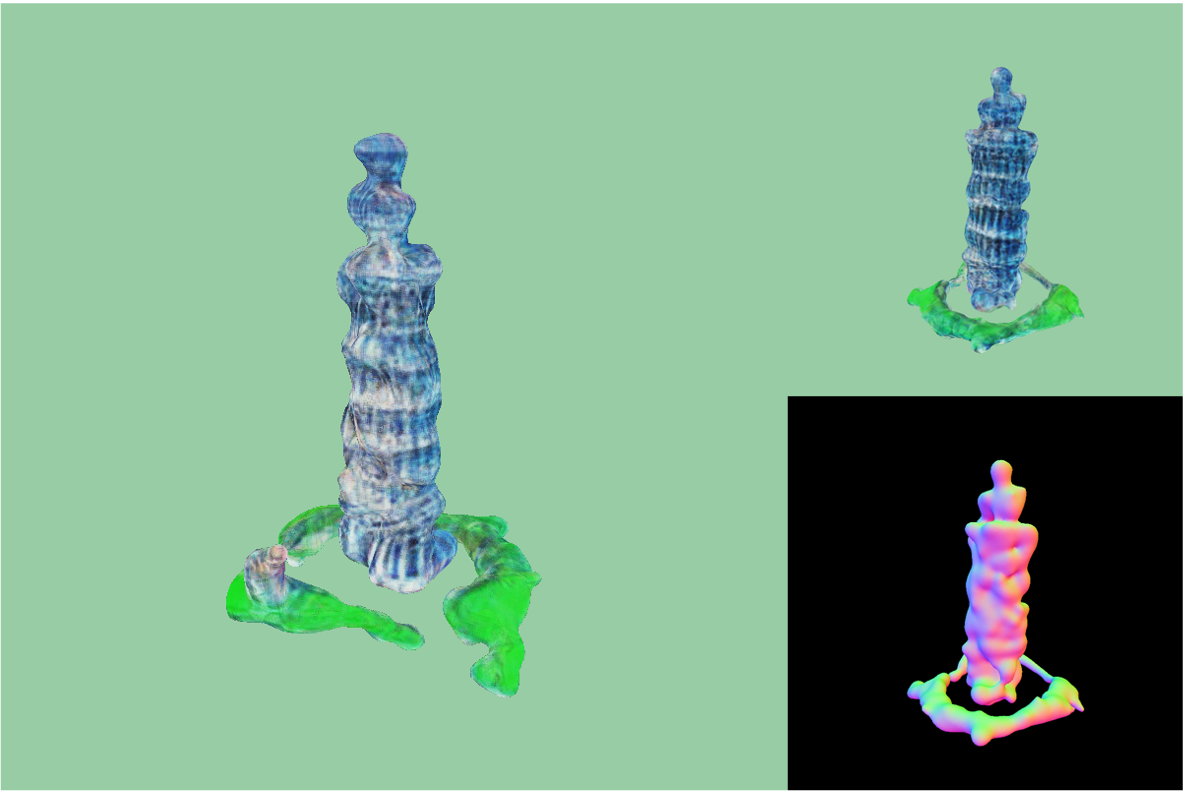}
        \end{minipage}
        &
         \begin{minipage}{.5\linewidth}
          \includegraphics[width=\linewidth,trim=1 1 1 1,clip]{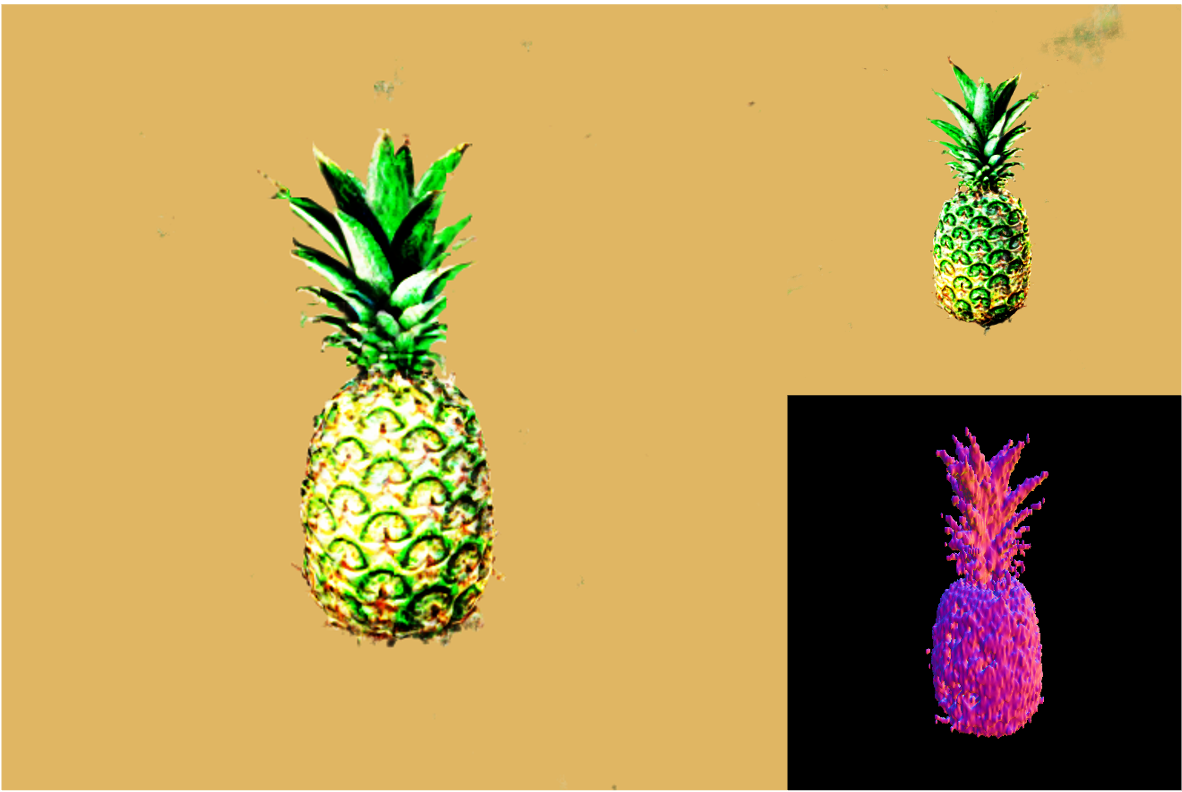}
        \end{minipage} \\
         & \cellcolor{Gainsboro!60}\textbf{2 minutes} & \cellcolor{Gainsboro!60}\textbf{2 hours 13 minutes} \\
      \parbox[t]{4mm}{\rotatebox[origin=c]{90}{+ \model}}  & \begin{minipage}{.5\linewidth}
          \includegraphics[width=\linewidth,trim=1 1 1 1,clip]{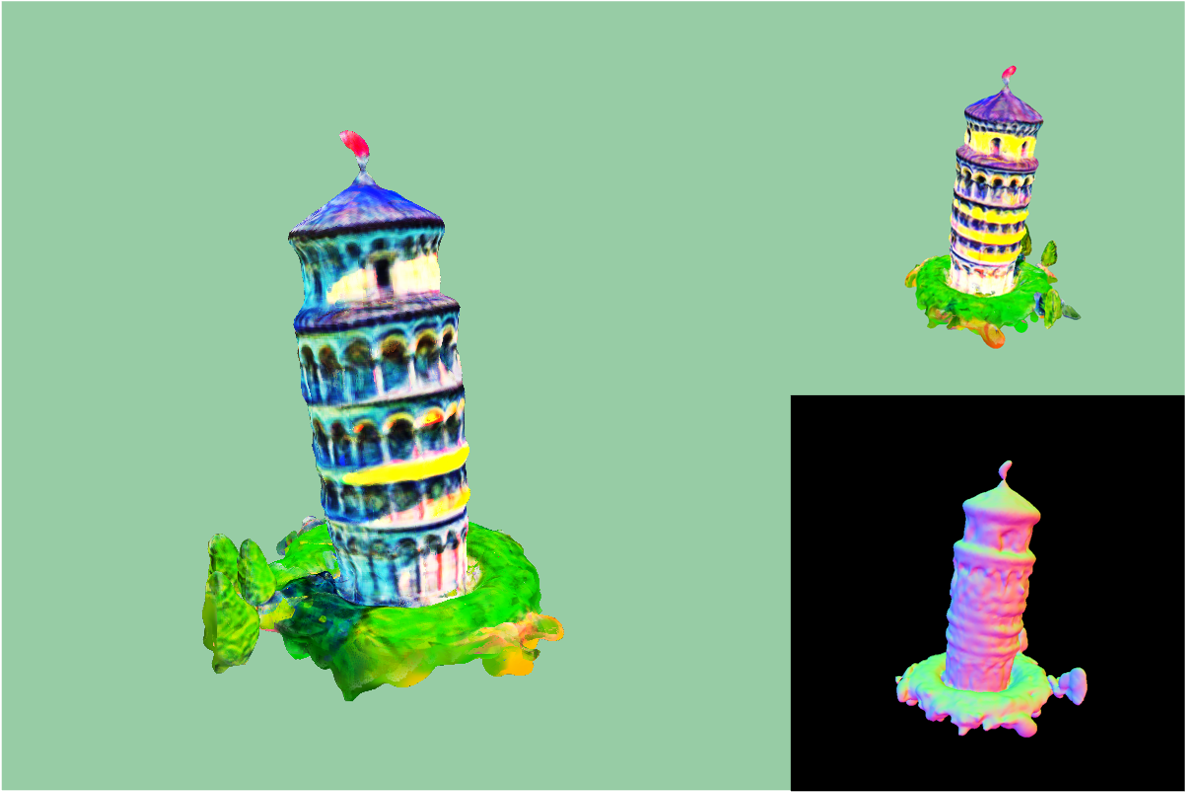}
        \end{minipage} &
         \begin{minipage}{.5\linewidth}
          \includegraphics[width=\linewidth,trim=1 1 1 1,clip]{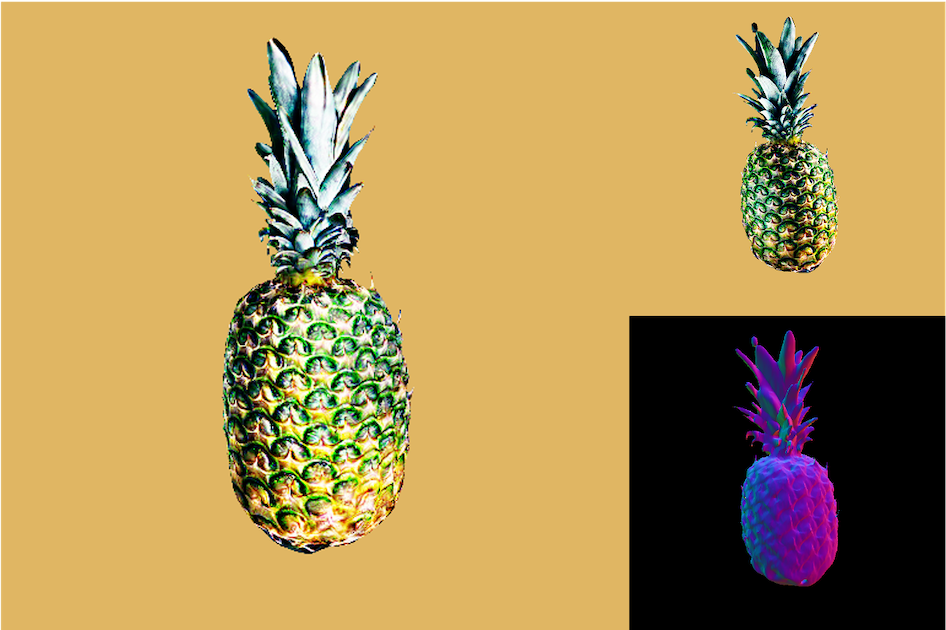}
        \end{minipage}  \\
         & \cellcolor{Gainsboro!60}\textbf{2 minutes} & \cellcolor{Gainsboro!60}\textbf{2 hours 13 minutes} \\
    & \textit{\thead{"the leaning tower of Pisa,\\ aerial view"}} & \textit{"a pineapple"} 
    \end{tabular}
\vspace{-2mm}
    \caption{We present two representative examples of applying \model. Gray rows denote runtime. Our framework trades parallel compute for speed and achieves more than 4x speedup when applied to both DreamGaussian~\citep{Tang_undated-od}\protect\footnotemark and ProlificDreamer~\citep{Wang2023-lp} while maintaining the generation quality. At the time when \model{} finishes, the baseline versions (Incomplete) exhibit significantly worse appearance and geometry.}
    \label{fig:teaser}
\vspace{-3mm}
\end{figure}

\footnotetext{We use batch size 16 for higher-quality generation, which requires longer time.}    
\section{Introduction}
\label{sec:intro}

Diffusion models~\citep{Sohl-Dickstein2015-sk, song2019generative, Ho2020-ki,Song2020-xi} 
have seen great success in a variety of domains,  including both 2D images~\citep{Rombach2022-uf,Meng2021-wh,Zhang2023-da,Zhang2023-da} and 3D shapes~\citep{Luo2021-do,Zhou2021-zo,Zhang2023-da,Zhang2023-da,Shue2022-dk}. Early attempts of applying diffusion frameworks to 3D shape generation require training separate models for each data category with 3D training samples. 
This problem is significantly alleviated by the use of large-scale text-conditioned 2D diffusion models~\citep{Rombach2022-uf}. Such large-scale foundation models, often trained on internet-scale data such as LAION-5B~\citep{Schuhmann2022-ke}, have exhibited remarkable semantic understanding of the visual world, which has inspired their use for creating 3D shapes in open-vocabulary settings. Among the most remarkable developments in recent years involves optimizing NeRF~\citep{Mildenhall2021-wv} scenes using Score Distillation Sampling (SDS)~\citep{Poole2022-sm} or Score Jacobian Chaining (SJC)~\citep{Wang2023-hq}, which push the 2D rendering of the 3D models to be more likely under the 2D diffusion prior through gradient-based optimization. More recent developments including Variational Score Distillation (VSD)~\citep{Wang2023-lp} further improve the quality of 3D shape generation. 


Despite promising results in generation quality, text-to-3D methods using SDS/VSD suffer from long generation time because gradient-based optimization is computationally expensive over the parameters of the 3D representation, and a large number of iterations are required for convergence.
Recent works~\citep{Tang_undated-od, Chen2023-dp, Yi2023-en} have explored using a novel efficient 3D representation, namely 3D Gaussian Splatting~\citep{Kerbl2023-kn}, to accelerate the distillation process. Others have explored amortizing the generation process by distilling a text-conditioned generator that can generate 3D shapes at test time in one step~\citep{Lorraine2023-fz}. Orthogonal to 
these methods, 
we propose a general
framework to accelerate text-to-3D creation (regardless of the underlying 3D representation) by leveraging parallel compute to perform the same number of optimization steps faster (in terms of wallclock time). 

Our method is based on a generalization of Picard iterations~\citep{Alfuraidan2016-nz} which we adapt to text-to-3D generation. 
While Picard iterations were previously shown to be effective in accelerating image generation from diffusion models through faster (parallel-in-time) ODE solving~\citep{Shih2023-nz}, we generalize the approach to more complex computation graphs involving multiple gradient updates and changes in the dimensions of the variables.
We show that all existing 3D representations for text-to-3D generation can be accelerated by our method, including the recently proposed 3D Gaussian Splatting.
Experimentally, we also achieve 4.7x times speedup with negligible degradation in generation quality.

\section{Related Works}
\label{sec:related}

\mypara{Text-to-Image diffusion models.} Diffusion models~\citep{Sohl-Dickstein2015-sk,song2019generative, Ho2020-ki, Song2020-xi} rose to popularity due to their superior performance and training stability compared to GANs~\citep{Goodfellow2014-ta}. They have been scaled up to large-scale foundation models, among the most popular of which are text-conditioned latent-diffusion models (LDM)~\citep{Rombach2022-uf,Saharia2022-hr,Balaji2022-nr}, which are built on the latent space of an autoencoder. 

Such text-conditional LDMs, often trained on internet-scale datasets~\citep{Schuhmann2022-ke}, can achieve high generation fidelity with reasonable faithfulness to input prompts. This has inspired many other remarkable applications such as adding additional control signals~\citep{Zhang2023-da}, customizing for subject-specific generation~\citep{Gal2022-ds,Ruiz2022-tx,Kumari2022-ou,Liu2023-wd}, and editing~\citep{Meng2021-wh, Mokady2022-jt}. The utilization of large-scale Text-to-Image diffusion models lies at the core of modern visual generative modeling, and is also central to our method for text-to-3D generation.


\mypara{Text/Image-to-3D generation via score distillation.} DreamFusion~\citep{Poole2022-sm} and Score Jacobian Chaining~\citep{Wang2023-hq} were among the first to propose lifting 2D diffusion models for 3D generation by Score Distillation Sampling (SDS), which propagates the score of pretrained diffusion models to the differentiable rendering of NeRF~\citep{Mildenhall2021-wv}. ProlificDreamer~\citep{Wang2023-lp} introduces Variational Score Distillation (VSD) which significantly improves the quality of 3D generation by adaptively training a LoRA model. Orthogonal to the distillation algorithm, other works such as~\citep{Lin2022-ha, Chen2023-sb} propose to optimize a more memory-efficient DMTet representation for high-resolution rendering. A more recent work~\citep{Tsalicoglou2023-js} proposes to volumetrically render signed distance fields (SDF) for better texture and mesh extraction. Others~\citep{Tang_undated-od, Yi2023-en, Chen2023-dp} have also investigated the use of 3D Gaussian Splatting~\cite{Kerbl2023-kn} as the underlying representation for fast and efficient generation, bringing the creation time down to as low as two minutes. 
Besides other applications such as controllable scene composition~\citep{Cohen-Bar2023-za,Bai2023-yp,Po2023-vc}, 
SDS is also widely applied in the Image-to-3D task, where Zero-1-to-3~\citep{Liu2023-kk} finetunes a view-dependent diffusion model for 3D generation given single images. Other works~\citep{Qian2023-rv,Liu2023-ut,Weng2023-db} further improved image-conditioned generation by integrating multi-view information and additional guidance into the generation process.

\mypara{Accelerating sampling with parallelism.}
Many works have studied accelerated sampling of generative models by leveraging parallel computation to trade compute for speed. For autoregressive models, prior works have used Jacobi/Gauss iteration~\citep{song2021accelerating} or predict/accept mechanisms~\citep{stern2018blockwise} to improve sampling speed. For diffusion models, parallelism based on fixed-point iterations has been shown to speed up sampling of pretrained diffusion models~\citep{Shih2023-nz}. Similarly, the goal of our work is to accelerate sampling, specifically for 3D generation models, which are known to have slow sampling speed. Our technique is most related to the  work of~\citet{Shih2023-nz}, which is inspired by the classic technique of Picard iterations to solve ODEs using parallel computation, allowing for the utilization of multiple GPUs to accelerate sampling. Unfortunately, the method of Picard iteration cannot be directly used to parallelize the sequential gradient update steps of 3D generation, because of the use of momentum-based gradient updates changes in dimension during optimization.

In this work, we seek to overcome the challenges of the representational differences of 3D generation and design an algorithm that accelerates the generation of all existing 3D representations. We do so by formulating Picard iterations~\citep{Shih2023-nz} for a wider family of sequential computation, enabling us to leverage parallel computational resources to accelerate 3D generation.

\section{Preliminary}
Recent Text-to-3D generation frameworks mostly rely on distilling knowledge from pretrained 2D diffusion models. We hereby introduce some backgrounds and notations helpful for our exposition.

\subsection{Text-to-3D Generation via Score Distillation}
2D diffusion models~\cite{Sohl-Dickstein2015-sk,song2019generative, Ho2020-ki, Song2020-xi} learn a distribution of images by adding noise to the ground-truth image $\rvx$ at noise level $t$, resulting in noisy images $\rvz_t$, and uses a score network $\eps_\phi(\rvz_t, t, y)$ with parameter $\phi$ and caption $y$ to estimate the gradient direction towards higher likelihood given $\rvz_t$. Such diffusion models can faithfully produce realistic images highly aligned with the input caption $y$. Their success has also inspired their use for 3D shape generation, which we shall introduce below. Suppose the 3D shape (\eg NeRF) parameterized by $\theta$, can be rendered into an image following a deterministic transformation $\bm{g}: \Theta\times \mathcal{C} \rightarrow \R^{H\times W\times C}$ which takes in shape parameter $\theta\in \Theta$ and camera parameters $c\in \mathcal{C}$. We seek an update rule for $\theta$ such that $\bm{g}(\theta, c)$ is a realistic image following caption $y$.

\mypara{Score Distillation Sampling.} The seminal works~\citep{Poole2022-sm,Wang2023-hq} on lifting 2D diffusion models for 3D creation propose to directly propagate diffusion score prediction towards NeRF parameters. The most prominent algorithm, Score Distillation Sampling (SDS), or Score Jacobian Chaining (SJC), states that a 3D model parameterized by 
can be guided to generate a scene with caption $y$ by following the update rule $\theta_{\tau+1} = \theta_\tau - \eta \gradnd{\mathcal{L}_{\text{SDS}}}{\theta_\tau}$. With $t\sim \mathcal{U}(0.02, 0.98)$, $\eps\sim \gN(\vzero, \mI)$, and $\rvz_t = \alpha_t\bm{g}(\theta, c) + \sigma_t\eps$,
\begin{align}
    \gradnd{\mathcal{L}_{\text{SDS}}}{\theta} = \E_{t,\eps}\Big[\omega(t)\Big(\eps_{\text{pretrain}}(\rvz_t, t, y) - \eps\Big)\parderiv{\bm{g}(\theta, c)}{\theta}\Big]
\end{align}
where $\omega(t)$ is a weighting function. In practice, the gradient update is done through Adam gradient update, which updates 3D models such that their rendering $\bm{g}(\theta, c)$ closely follows distributions of the pretrained 2D diffusion prior.

\mypara{Variational Score Distillation.} Despite success in zero-shot 3D NeRF generation, SDS often suffers from over-saturation and simplistic geometry. To enhance the generative quality, another recent work~\citep{Wang2023-lp} proposes Variational Score Distillation (VSD), which replaces the noise sample $\eps$ with a trainable LoRA diffusion with parameter $\phi$ such that
\begin{align}
\begin{split}
    \gradnd{\mathcal{L}_{\text{VSD}}}{\theta} = \E_{t,\eps}\Big[\omega(t)\Big(&\eps_{\text{pretrain}}(\rvz_t, t, y) - \\&\eps_\phi(\rvz_t, t, c, y)\Big)\parderiv{\bm{g}(\theta, c)}{\theta}\Big]
\end{split}
\end{align}
where the LoRA model is adaptively trained to fit the distribution of the current render $\bm{g}(\theta, c)$.


\mypara{Generative 3D Gaussian Splatting.} 3D Gaussian Splatting~\citep{Kerbl2023-kn} is a recently developed 3D representation for efficient rendering. It is represented by a set of 3D Gaussians which are optimized with differentiable rasterizers. More recent works~\citep{Tang_undated-od, Yi2023-en, Chen2023-dp} have adopted this representation for score distillation, which greatly reduces the generation time. However, this representation is different from others in that the number of Gaussians can change during the optimization process due to its split-and-prune operations. 

\subsection{Picard Iterations}

\begin{figure}[t]
    \centering
    \includegraphics[width=0.8\linewidth]{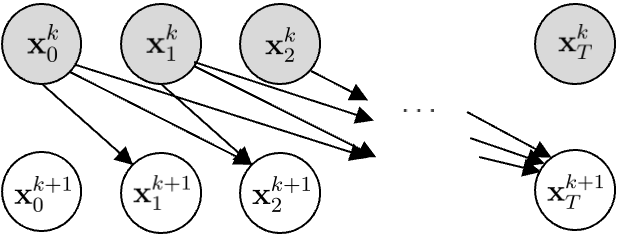}
    \caption{Picard dependency graph. Gray nodes have outgoing edges to all subsequent nodes in $k+1$-th iteration and are independent of each other. This allows parallel computation of $s(\rvx_\tau^k, \tau)$ for all $\tau \in [0,T-1]$.}
    \label{fig:picard-dependency}
\end{figure}

The classic Picard iteration approximates the solution trajectory  of an ODE $\rvx_{0:\tau}$ ending at time $\tau$ 
\begin{align}\label{eq:int}
    \rvx_\tau = \rvx_0 + \int_{0}^{\tau} s(\rvx_u, u) \dd u
\end{align} 
via fixed-point iteration, by starting with a guess of the full trajectory $\rvx^{k=0}_{0:T}$ and iteratively refining until convergence using the following iteration rule for each iteration $k$:
\begin{align}\label{eq:picard}
    \rvx_\tau^{k} = \rvx_0^{k-1} + \int_{0}^{\tau} s(\rvx_u^{k-1} , u) \dd u
\end{align} 
which under mild conditions will converge to a unique ODE solution given initial condition $\rvx_0$. The discrete-time approximation is
\begin{align}\label{eq:discrete-picard}
    \rvx_\tau^{k} = \rvx_0^{k-1} + \frac{1}{T}\sum_{i=0}^{\tau-1} s(\rvx_i^{k-1} , \frac{i}{T})
\end{align} 
which is guaranteed to converge in $T$ steps, but in practice often converges much faster in $K \ll T$ iterations.
This procedure allows us to leverage parallel computation to converge to the solution in $O(K)$ steps. In \figref{fig:picard-dependency} we depict the computation graph of Picard iterations, where we see long-range dependencies allowing for information to propagate quickly to the end of the sequence.

Previously, \citet{Shih2023-nz} leveraged Picard iterations to accelerate sampling for diffusion models on images. 
Once converged, we simply extract the endpoint of the solution trajectory of the final iteration $\rvx_{\tau=T}^{k=K}$ as the sample of the diffusion model. 

In this work, we seek to extend the method of Picard iterations to Text-to-3D generation via score distillation. However, generalizing this process to 3D generation is non-trivial because, unlike~\citet{Shih2023-nz}, solution trajectories for 3D generation is not in data space, but rather in ``parameter'' space. This presents key challenges: 1) for methods such as 3D Gaussian Splatting, the parameter space has changing dimensionality over the length of the trajectory, 2) all score distillation methods involve sequential gradient update using Adam optimizer, as opposed to a trivial prefix sum. 
Therefore, we must investigate more deeply to come up with a generalized formulation of Picard iteration that is applicable to 3D generation.

\section{Method}

\begin{figure*}
    \centering
    \includegraphics[width=0.9\linewidth]{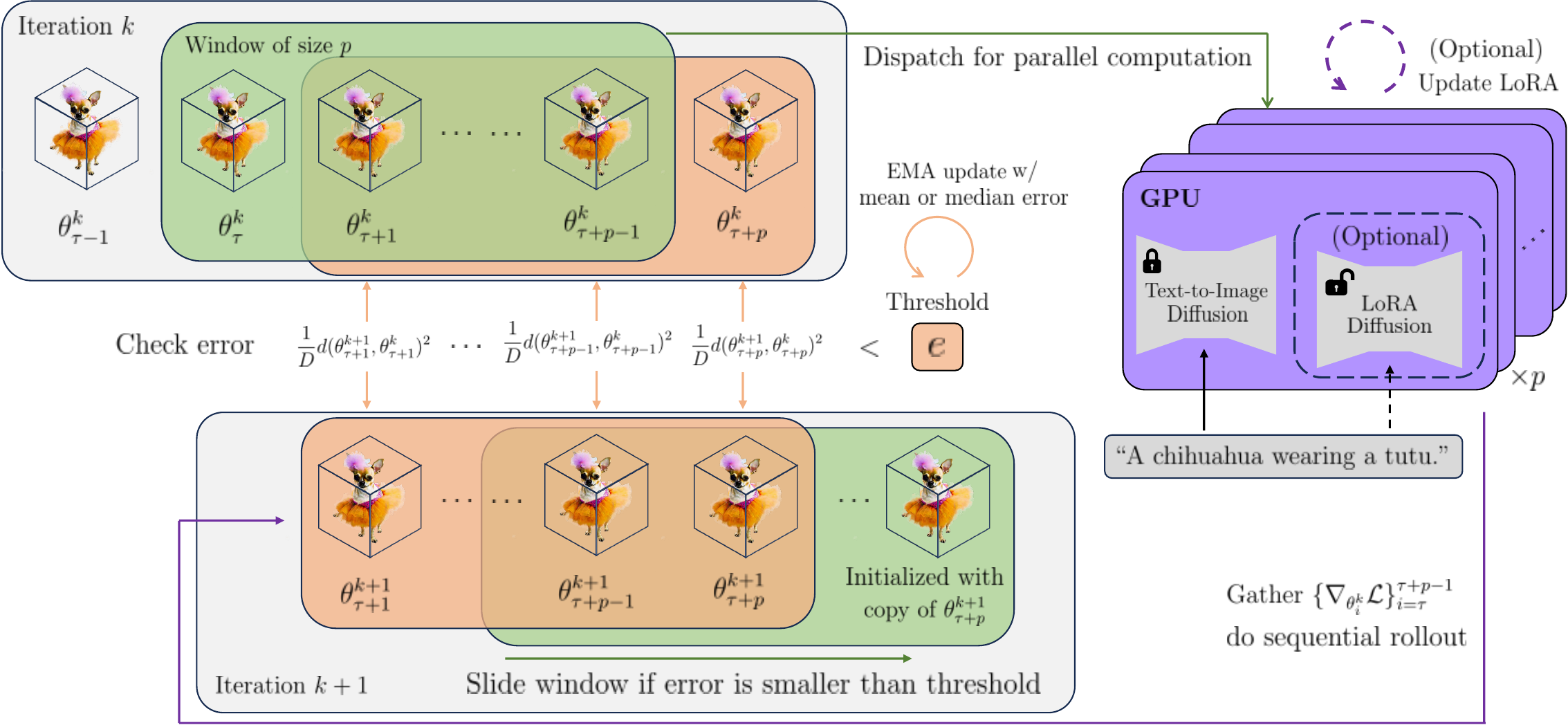}
    \caption{Overview of \model. Starting from top left, for iteration $k$, we initialize a window of 3D shapes (in green) with dimension $D$ and dispatch them to $p$ GPUs for parallelly computing the SDS/VSD gradients, which are gathered for rollout using the rule in \equref{eq:general-rollout}. The resulting shapes (in orange) for iteration $k+1$ are compared to those in iteration $k$. The window is slid forward until the error at that time step is not smaller than the threshold $e$, which is adaptively updated with the mean/median error of the window. Optionally, in the case of VSD, we keep independent copies of LoRA diffusion on all GPUs which are updated independently without extra communication.}
    \label{fig:main}
\end{figure*}

Despite the impressive quality of the generation results from VSD and SDS, all of score distillation methods suffer from long generation time (\eg ~10 hours for one generation from VSD on an NVIDIA A100), making them prohibitively expensive to use in practice. Motivated by this problem, we seek to design an acceleration algorithm for all existing 3D representations suitable for score distillation.

\subsection{Score Distillation as Sequential Computation}

Our key insight is that the parameter update rules for SDS/VSD can roughly be written as a sequential computation $\theta_{\tau+1} = \theta_\tau + \eta \gradnd{\mathcal{L}}{\theta_\tau}$ of a form that is similar to Picard iterations, where $\mathcal{L}$ can either be $\mathcal{L}_{\text{SDS}}$ or $\mathcal{L}_{\text{VSD}}$ and $\eta$ is step size. However, we cannot directly apply Picard iterations because these methods have further complications in their sequential computations. 
For example, SDS/VSD rely on momentum-based optimizers such as Adam~\citep{Kingma2014-mf} to perform the parameter update, where the momentum prevents us from directly using vanilla Picard iteration. Special representations such as Gaussian Splatting intertwine gradient updates with splitting operations, which increase the dimensionality of $\theta$, a property that similarly prevents the na\"ive use of Picard.


\if\arxiv0
\begin{table*}[t]
    \centering
    \begin{tabular}{ccc}
    \toprule
      & Picard  & Generalized \\
      \midrule
       Sequential output $\theta_{\tau+1}^{k+1}$  & $g(\theta_\tau^k)$ &  $g(\theta_\tau^k)$ \\
       Update function $s(\theta_\tau^k)$ & $s(\theta_\tau^k) =\frac{1}{\eta}(g(\theta_\tau^k)- \theta_\tau^k )$  & $s( \theta_\tau^k) = h(g(\theta_\tau^k), \theta_\tau^k )$ \\
       Output parameterization & $\theta_{\tau+1}^{k+1} =\theta_\tau^k + \eta s( \theta_\tau^k)$ & $ \theta_{\tau+1}^{k+1} = h^\dagger(s(\theta_\tau^k), \theta_\tau^k )$ \\
       Iteration rule & $\theta_\tau^k = \theta_0^{k-1} + \eta \sum_{i=0}^{\tau-1} s(\theta_i^{k-1})$ & $\theta_{\tau}^k = h^\dagger(s(\theta_{\tau-1}^{k-1}),  \dots h^\dagger(s(\theta_0^{k-1}), \theta_0^{k-1}) \dots )$\\
    \bottomrule
    \end{tabular}
    \caption{Comparison between Picard and generalized iterations.}
    \label{tab:comp}
\end{table*}
\fi

\subsection{Generalizing Picard Iterations}

We now present a generalized version of Picard iterations that will enable us to apply the same parallelization techniques to more complicated computation graphs, encompassing cases such as Gaussian Splatting and SDS/VSD.

In standard Picard iterations, our goal is to parallelize an ODE that takes additive sequential updates of the form 
\begin{align}
    \theta_{\tau+1} = g(\theta_{\tau}) = \theta_{\tau} + \eta s( \theta_{\tau})
\end{align}
where $s(\cdot)$ is the drift function of the ODE, and the eventual \textit{computational unit} that we will parallelize over. To take the first step towards generalizing Picard iteration, we rearrange to write this computational unit $s$ explicitly:
\begin{align}
    s( \theta_{\tau} ) = \frac{1}{\eta}(g(\theta_{\tau})- \theta_{\tau})
\end{align}
where $g(\cdot)$ denotes the underlying function outputting the next parameter $\theta_{\tau}$.
Choosing the computational unit $s$ as the drift is natural for ODEs, and is convenient because unrolling drifts can be easily done via a summation. However, this choice of computing the drift cannot be generalized to settings such as expanding/shrinking of dimensions and discrete domains, where subtraction operator can be unnatural or even undefined. Moreover, in settings using momentum-based updates (Adam), applying drift-based error accumulation to the momentum terms can lead to very poor performance.

To generalize Picard iterations, our insight is that we can consider many different choices of the computational unit $s$. 
We can write $s$ as some general function of $g(\theta_\tau)$ and $\theta_\tau$:
\begin{align}
    s( \theta_\tau) = h(g(\theta_\tau), \theta_\tau )
\end{align}
where $h$ is equipped with $h^\dagger(\cdot,\cdot)$, a pseudo-inverse of $h$ \wrt the first argument $g(\theta_\tau)$. 
Under this condition, we can write the following iteration rule:
\begin{align}\label{eq:general-rollout}
    \theta_{\tau}^{k} = h^\dagger\left(s(\theta_{\tau-1}^{k-1}), h^\dagger\left(s(\theta_{\tau-2}^{k-1}), \dots h^\dagger(s(\theta_{0}^{k-1}), \theta_0^{k-1})\right) \right)
\end{align}
This iteration rule can be understood as the generalized form of Picard iterations. Similar to Picard iterations, we 1) perform the computation units $s(\theta^{k-1}_{0:T})$ in parallel and 2) do a sequential unrolling of the trajectory from iteration $k-1$ to arrive at iteration $k$. In contrast, Picard iteration has a simple form for the sequential unrolling, \ie cumulative sum (as in \tabref{tab:comp}). 


We further show in \appref{app:proof} that the existence of $h^\dagger$ paired with the iteration rule in~\cref{eq:general-rollout} is well-defined and guarantees a fixed-point solution. 


Finally, we proceed by demonstrating two concrete choices of $s$ for generalized Picard iterations on 3D Gaussian Splatting and SDS, both of which pose problems for naive Picard iteration.



\begin{algorithm}[t]
\caption{Generalized Picard Iteration}\label{alg:algo-rough}
\begin{algorithmic}
\algtext*{EndWhile} 
\algtext*{EndFor}   
\State {\bf Input:} Initial parameter $\theta$, pseudo-inverse $h^\dagger$, drift $s$, maximum time $T$
\State {\bf Output:} Score distillation output 
\State $\theta_\tau^0 \leftarrow \theta, \quad \forall \tau\in [0, T]$
\State $\tau,k \leftarrow 0,0$
\While{not converged}
\For{$\tau$ from $0$ to $T-1$} compute $s(\theta_{\tau}^k)$ in parallel
\EndFor
\For{$\tau$ from $0$ to $T-1$} $\theta_{\tau+1}^{k+1} \leftarrow h^\dagger(s(\theta_{\tau}^k), \theta_{\tau}^k)$
\EndFor
\State $k\leftarrow k+1$
\EndWhile

\State \Return $\theta_{T}^k$
\end{algorithmic}
\end{algorithm}

\mypara{Example 1: 3D Gaussian Splatting}

Recently, 3D Gaussian Splatting~\citep{Kerbl2023-kn} has been adopted for 3D generative models using SDS~\citep{Tang_undated-od, Yi2023-en, Chen2023-dp}. A unique property of this representation is its ability to dynamically change its dimension (\ie number of Gaussians) during the optimization process. This poses significant challenges to the classical Picard iterations as it assumes fixed dimensions throughout (\ie $\theta_\tau,\theta_{\tau+1}$ have the same dimension).

Concretely, in the case of 3D Gaussian Splatting where the number of points may increase, the difference in dimension $\theta_{\tau}^k$ and $g(\theta_{\tau}^k)$ can be addressed by designing $h(g(\theta_{\tau}^k),\theta_{\tau}^k)$ to be
\begin{align}
    s( \theta_\tau^k) = \frac{1}{\eta}(\text{proj}(g(\theta_\tau^k))- \theta_\tau^k )
\end{align}
where $\text{proj}: \R^N\rightarrow \R^M$, $N\ge M$, is a projection function from $\R^N$ to $\R^M$ by deleting a subset of points from its input. Its pseudo-inverse can be 
\begin{align}
    \theta_{\tau+1}^{k+1} = \text{unproj}(\theta_\tau^k + \eta s( \theta_\tau^k))
\end{align}
where $\text{unproj}: \R^M\rightarrow \R^N$ is a dimensionality-increasing function that adds new points to the current parameter. Following \citet{Tang_undated-od}, we use the split-and-clone function as our $\text{unproj}(\cdot)$ function. 

\if\arxiv1
\begin{table*}[t]
    \centering
    \begin{tabular}{ccc}
    \toprule
      & Picard  & Generalized \\
      \midrule
       Sequential output $\theta_{\tau+1}^{k+1}$  & $g(\theta_\tau^k)$ &  $g(\theta_\tau^k)$ \\
       Update function $s(\theta_\tau^k)$ & $s(\theta_\tau^k) =\frac{1}{\eta}(g(\theta_\tau^k)- \theta_\tau^k )$  & $s( \theta_\tau^k) = h(g(\theta_\tau^k), \theta_\tau^k )$ \\
       Output parameterization & $\theta_{\tau+1}^{k+1} =\theta_\tau^k + \eta s( \theta_\tau^k)$ & $ \theta_{\tau+1}^{k+1} = h^\dagger(s(\theta_\tau^k), \theta_\tau^k )$ \\
       Iteration rule & $\theta_\tau^k = \theta_0^{k-1} + \eta \sum_{i=0}^{\tau-1} s(\theta_i^{k-1})$ & $\theta_{\tau}^k = h^\dagger(s(\theta_{\tau-1}^{k-1}),  \dots h^\dagger(s(\theta_0^{k-1}), \theta_0^{k-1}) \dots )$\\
    \bottomrule
    \end{tabular}
    \caption{Comparison between Picard and generalized iterations.}
    \label{tab:comp}
    \vspace{-5mm}
\end{table*}
\fi


\mypara{Example 2: Adam gradient updates}

With methods that use optimizers such as Adam, the sequential update involves a momentum term (here denoted as $m_\tau^k$) which necessitates the use of the generalized form of Picard iterations, since applying vanilla Picard iterations on $s((\theta_\tau^k, m_\tau^k))$ will lead to poor updating of the additional momentum parameters.


To incorporate the additional momentum update rules, we can simply design $h(g(\theta_{\tau}^k, m_{\tau}^k),(\theta_{\tau}^k, m_{\tau}^k))$ to be $\gradnd {\mathcal{L}}{\theta_{\tau}^k}$, which gives us (assuming learning rate $\eta$) the natural pseudo-inverse
\begin{align}
    \theta_{\tau+1}^{k+1}, m_{\tau+1}^{k+1} = \text{Adam}( s((\theta_{\tau}^k, m_{\tau}^k)), (\theta_{\tau}^k, m_{\tau}^k) )
\end{align}
and the generalized Picard update becomes
\begin{align}
    \theta_{\tau+1}^{k+1}, m_{\tau+1}^{k+1} \gets g(\theta_{\tau}^k, m_{\tau}^k) = \text{Adam}(\gradnd {\mathcal{L}}{\theta_{\tau}^k}, (\theta_{\tau}^k, m_{\tau}^k)  )
\end{align}



For clarity, we additionally provide a high-level algorithm in \algoref{alg:algo-rough}. Because of parallel computation of $s(\cdot)$, this algorithm can converge in $\mO(K)$ where $K\ll T$.

\subsection{Practical Decisions}

We note several practical considerations that are significant for empirical success.

\mypara{Sliding window.} Although one can in theory parallelize the entire trajectory, it is impractical to start by keeping all of $\{\theta_\tau^0\}_{\tau=0}^{T}$ in memory. In 3D generation settings, $T$ can usually be on the order of 10K and each parameter can cost large amounts of memory. We therefore similarly take inspiration from~\citep{Shih2023-nz} and employ a batched window scheme such that the Picard iteration is only performed on $\theta_{\tau:\tau+p}^{k}$ and the window is slid across until the fixed-point convergence error at the starting time step is above a threshold $e$. More details on the distance metric and how to handle dimension mismatch can be found in \appref{app:practice}.

In addition, to maximize efficiency, we want to parallelize the expensive calculation of SDS/VSD gradients, so we need to put one pretrained diffusion model on each GPU. We set the window size to be one less than the total number of GPUs, and use the remaining one for sequential rollout.

\mypara{Eliminating stochasticity.} As Picard iterations' convergence depends on the deterministic fixed-point iteration scheme, which requires deterministic gradient for the same $\theta_\tau^k$ at each $\tau$. However, the calculation of both $\gradnd{\mathcal{L}_{\text{SDS}}}{\theta_\tau}$ and $\gradnd{\mathcal{L}_{\text{VSD}}}{\theta_\tau}$ are stochastic due to Monte Carlo approximation of the expectation. To resolve this, we simply fix the random seed for each iteration, which works well empirically.

\mypara{Parallelizing Variational Score Distillation.} Variational Score Distillation requires training an additional LoRA model to adapt to the distribution of the current generated results. There are two possible solutions for parallelization: (1) we can update LoRA parameters similarly as our 3D parameters, by calculating their gradients on separate GPUs and aggregating them on the remaining GPU; (2) we can keep different LoRA models on different GPUs and separately update each without passing them back for aggregation. The first approach results in more accurate gradients as it seeks to parallelize updates of a single LoRA model, but we find that passing around LoRA parameters across GPUs is very expensive and can undermine any speed gain for the actual 3D model updates. We therefore avoid this approach. The second solution provides less accurate LoRA gradients in theory because our 3D model parameters can be passed to different LoRA models on different GPUs at any iteration, which can provide different LoRA updates at any point in time. However, we observe that since each of our 3D model parameters in the window is randomly allocated to different GPUs for gradient calculation, there is approximately equal probability that each LoRA model will observe all models in the window. This means, roughly speaking, each LoRA model will learn the distribution of all 3D models in the current window. Therefore, updating LoRA separately on different GPUs gives valid guidance and eliminates the need for communicating additional parameters across GPUs.

\mypara{Adaptive threshold.} 
Fixed-point errors control how close one is to the true trajectory. Smaller thresholds lead to slow convergence and larger thresholds lead to worse generation quality. In practice, we observe the threshold can be different for different prompts and 3D representations. To avoid the need for excessively tuning thresholds, we propose to adaptively update the threshold by the exponential moving average (EMA) of the mean or median error of the current window. With EMA decay rate $\gamma$, window of size $p$ starting at time $\tau$ and iteration $k$, and assume the parameter has dimension $D$, we update threshold $e$ by
\begin{align}
    e\leftarrow \gamma e + (1 - \gamma) * M(\{ \frac{1}{D}d(\theta_{\tau+i}^{k+1}, \theta_{\tau+i}^k)^2\}_{i=1}^p)
\end{align} where $M$ is a mean or median function.
We investigate the effect of the EMA parameters in ablation studies.



Our final method is depicted in~\figref{fig:main}, and we provide a detailed practical algorithm in \appref{app:algo}.

\begin{figure*}[th]
    \centering
     \begin{tabular}{@{}>{\kern-\tabcolsep}c@{}c@{}c@{}c<{\kern-\tabcolsep}@{}}
        DreamFusion~\citep{Poole2022-sm}  & + \model &DreamFusion~\citep{Poole2022-sm} & + \model \\
         \begin{minipage}{.25\textwidth}
          \includegraphics[width=\linewidth,trim=1 1 1 1,clip]{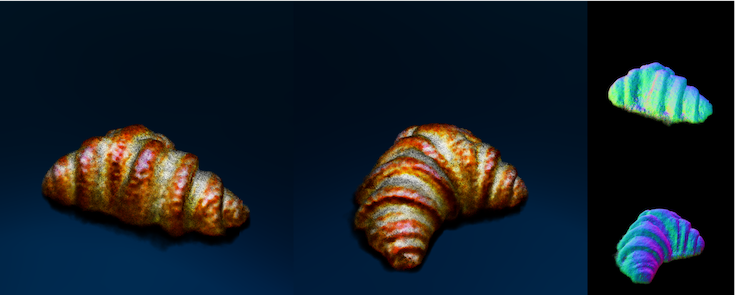}
        \end{minipage} & 
        \begin{minipage}{.25\textwidth}
          \includegraphics[width=\linewidth,trim=1 1 1 1,clip]{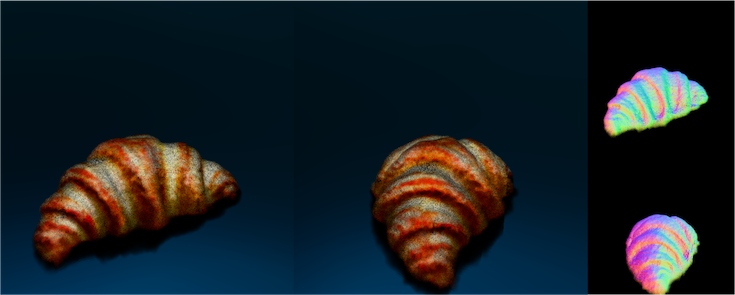}\end{minipage}  &
          \begin{minipage}{.25\textwidth}
          \includegraphics[width=\linewidth,trim=1 1 1 1,clip]{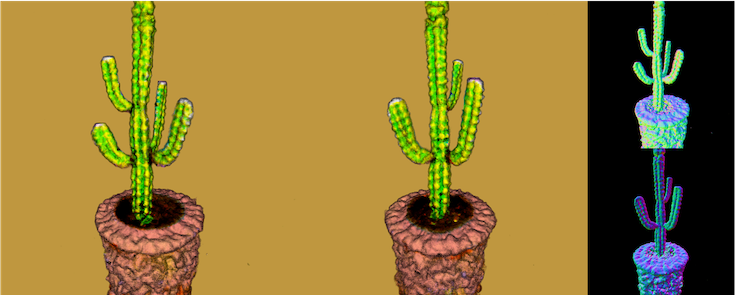}\end{minipage}   & 
          \begin{minipage}{.25\textwidth}
          \includegraphics[width=\linewidth,trim=1 1 1 1,clip]{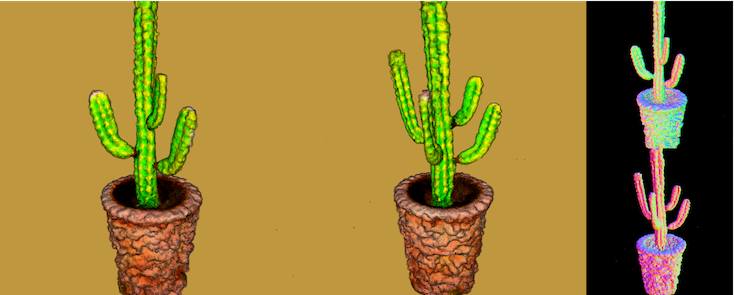} \end{minipage}  \\
          \rowcolor{Gainsboro!60}
        2 hours 10 minutes & \textbf{32 minutes} & 2 hours 9 minutes & \textbf{31 minutes} \\
        \multicolumn{2}{c}{\textit{"a delicious croissant"
}} & \multicolumn{2}{c}{\textit{"a small saguaro cactus planted in a clay"}} \\
Magic3D~\citep{Lin2022-ha}  & + \model &Magic3D~\citep{Lin2022-ha} & + \model \\
         \begin{minipage}{.25\textwidth}
          \includegraphics[width=\linewidth,trim=1 1 1 1,clip]{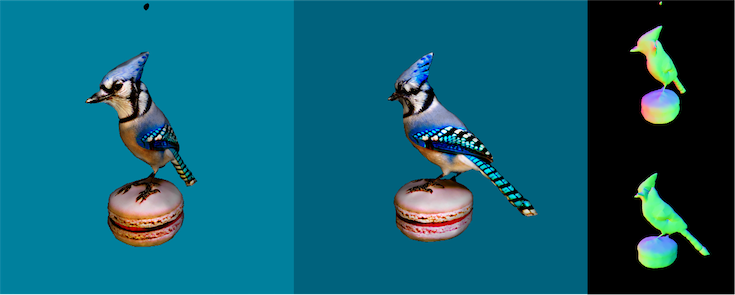}
        \end{minipage} & 
        \begin{minipage}{.25\textwidth}
          \includegraphics[width=\linewidth,trim=1 1 1 1,clip]{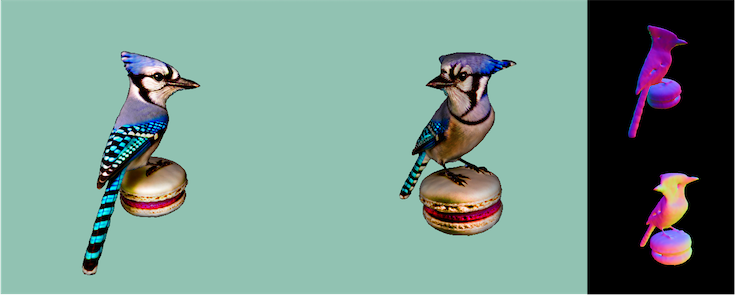}\end{minipage}  &
          \begin{minipage}{.25\textwidth}
          \includegraphics[width=\linewidth,trim=1 1 1 1,clip]{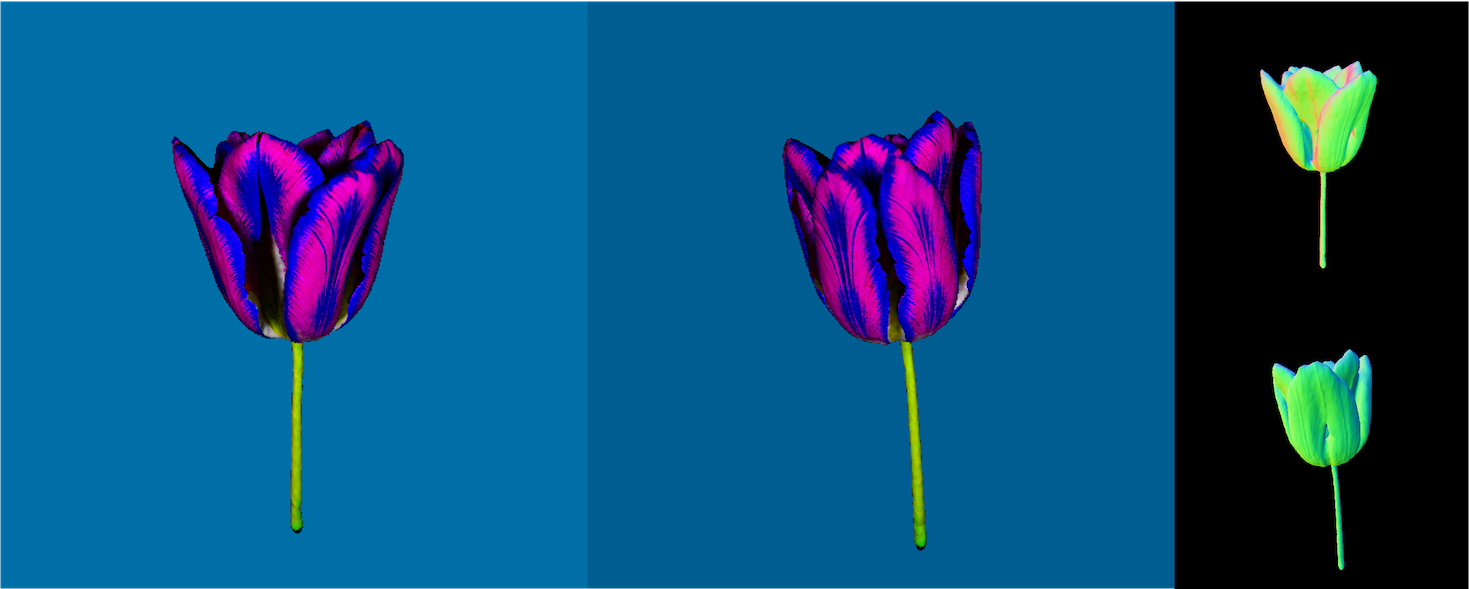}\end{minipage}   & 
          \begin{minipage}{.25\textwidth}
          \includegraphics[width=\linewidth,trim=1 1 1 1,clip]{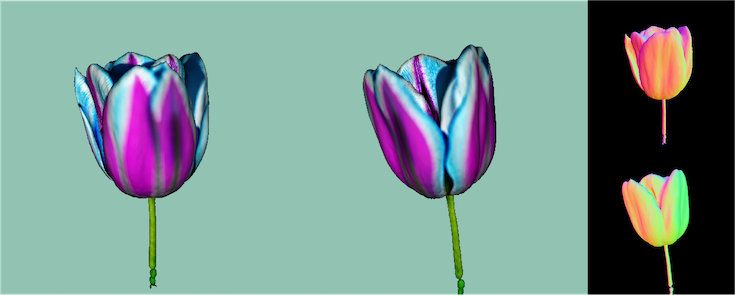} \end{minipage}  \\
          \rowcolor{Gainsboro!60}
        3 hours 34 minutes & \textbf{52 minutes} & 3 hours 35 minutes & \textbf{54 minutes} \\
        \multicolumn{2}{c}{\textit{"a DSLR photo of a blue jay standing on a macaron"
}} & \multicolumn{2}{c}{\textit{"a DSLR photo of a blue tulip"}} \\
        
TextMesh~\citep{Tsalicoglou2023-js}  & + \model &TextMesh~\citep{Tsalicoglou2023-js} & + \model \\
         \begin{minipage}{.25\textwidth}
          \includegraphics[width=\linewidth,trim=1 1 1 1,clip]{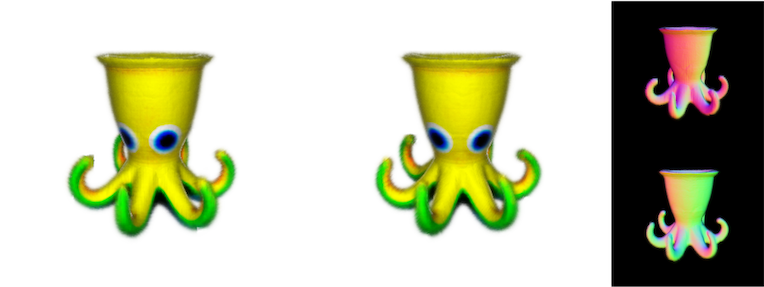}
        \end{minipage} & 
        \begin{minipage}{.25\textwidth}
          \includegraphics[width=\linewidth,trim=1 1 1 1,clip]{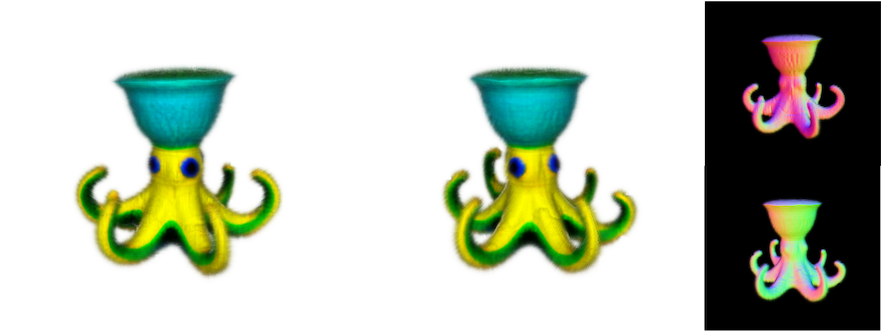}\end{minipage}  &
          \begin{minipage}{.25\textwidth}
          \includegraphics[width=\linewidth,trim=1 1 1 1,clip]{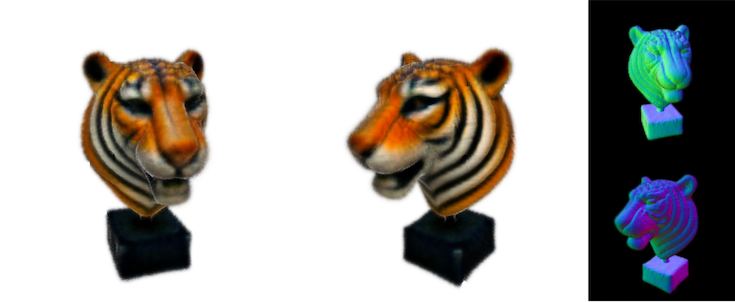}\end{minipage}   & 
          \begin{minipage}{.25\textwidth}
          \includegraphics[width=\linewidth,trim=1 1 1 1,clip]{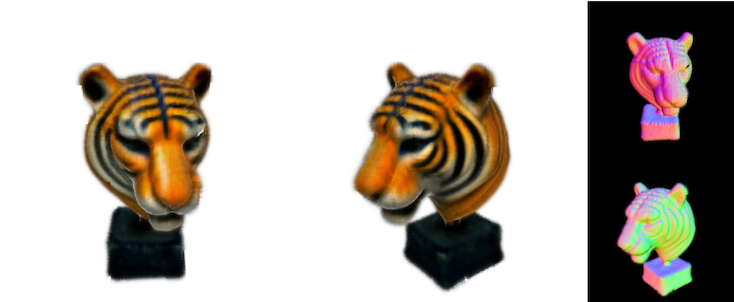} \end{minipage}  \\
          \rowcolor{Gainsboro!60}
        2 hours 5 minutes & \textbf{30 minutes} & 2 hours 6 minutes & \textbf{30 minutes} \\
        \multicolumn{2}{c}{\textit{\thead{"a ceramic upside down yellow octopus\\ holding a blue-green ceramic cup"}
}} & \multicolumn{2}{c}{\textit{"a highly detailed stone bust of the tiger"}} \\

DreamGaussian~\citep{Tang_undated-od}  & + \model &DreamGaussian~\citep{Tang_undated-od} & + \model \\
         \begin{minipage}{.25\textwidth}
          \includegraphics[width=\linewidth,trim=1 1 1 1,clip]{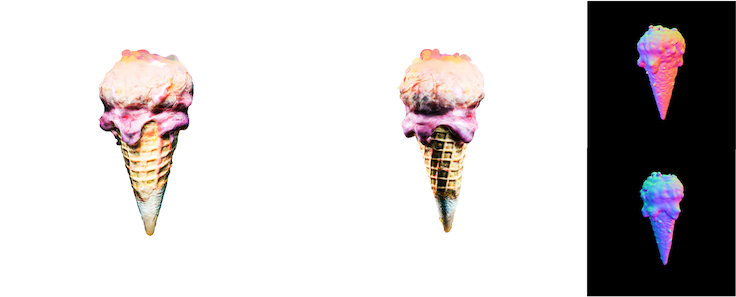}
        \end{minipage} & 
        \begin{minipage}{.25\textwidth}
          \includegraphics[width=\linewidth,trim=1 1 1 1,clip]{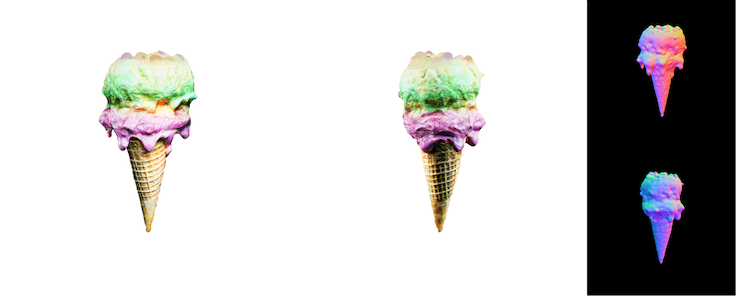}\end{minipage}  &
          \begin{minipage}{.25\textwidth}
          \includegraphics[width=\linewidth,trim=1 1 1 1,clip]{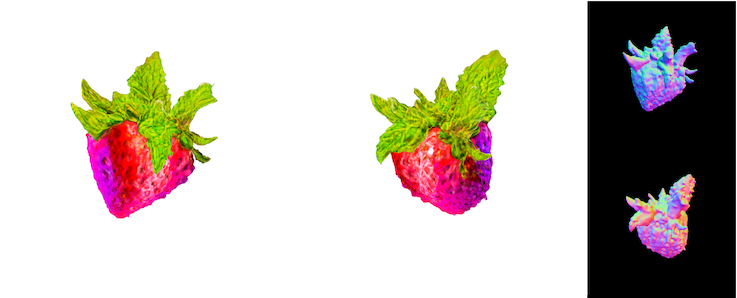}\end{minipage}   & 
          \begin{minipage}{.25\textwidth}
          \includegraphics[width=\linewidth,trim=1 1 1 1,clip]{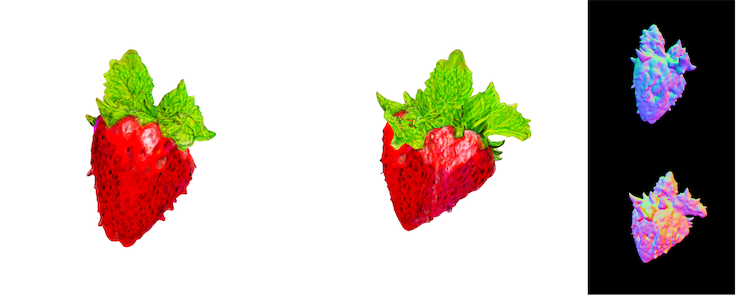} \end{minipage}  \\
          \rowcolor{Gainsboro!60}
        11 minutes & \textbf{2 minutes} & 11 minutes& \textbf{2 minutes} \\
        \multicolumn{2}{c}{\textit{"a photo of an ice cream
"
}} & \multicolumn{2}{c}{\textit{"a ripe strawberry
"}} \\

ProlificDreamer~\citep{Wang2023-lp}  & + \model &ProlificDreamer~\citep{Wang2023-lp} & + \model \\
         \begin{minipage}{.25\textwidth}
          \includegraphics[width=\linewidth,trim=1 1 1 1,clip]{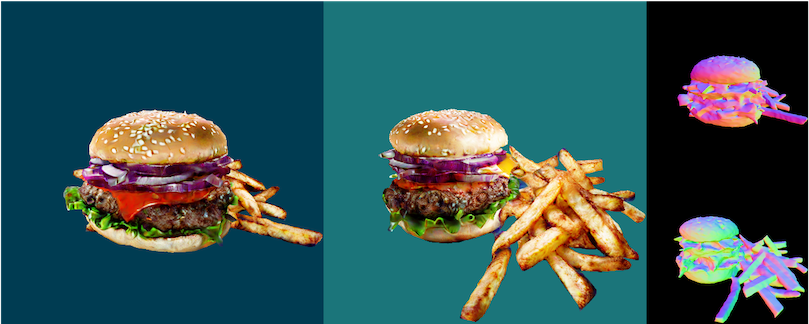}
        \end{minipage} & 
        \begin{minipage}{.25\textwidth}
          \includegraphics[width=\linewidth,trim=1 1 1 1,clip]{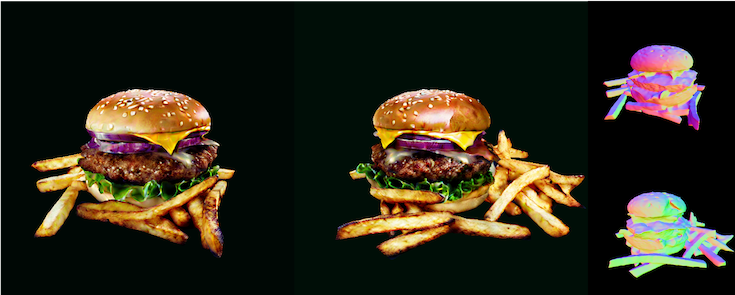}\end{minipage}  &
          \begin{minipage}{.25\textwidth}
          \includegraphics[width=\linewidth,trim=1 1 1 1,clip]{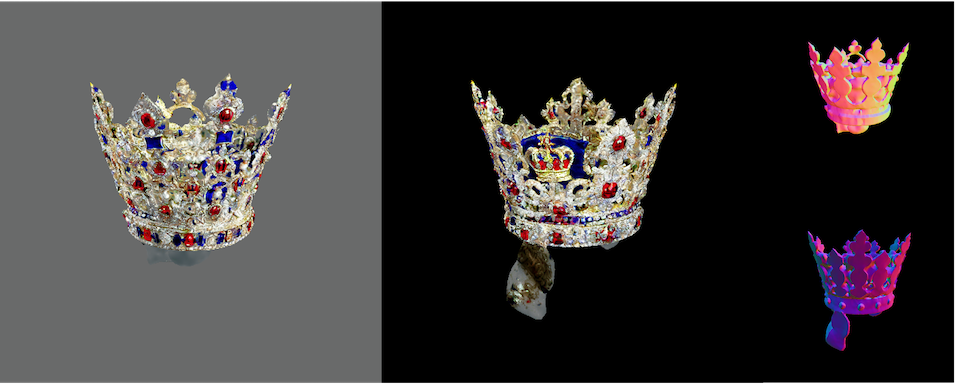}\end{minipage}   & 
          \begin{minipage}{.25\textwidth}
          \includegraphics[width=\linewidth,trim=1 1 1 1,clip]{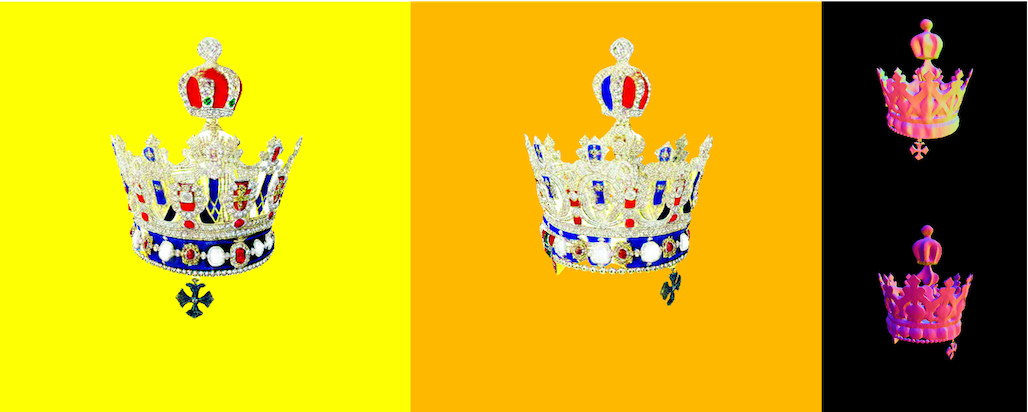} \end{minipage}  \\
          \rowcolor{Gainsboro!60}
        9 hours 21 minutes & \textbf{2 hours 10 minutes} & 9 hours 13 minutes & \textbf{2 hours 15 minutes} \\
        \multicolumn{2}{c}{\textit{"fries and a hamburger"
}} & \multicolumn{2}{c}{\textit{"an imperial state crown of England
"}} \\
         
     \end{tabular}
    \caption{Visual comparisons. Methods using \model{} achieve equally high-quality generation with a much shorter runtime. }
    \label{fig:text-to-3d}
\end{figure*}

\section{Experiments}

Running \model{} for $T$ iterations is guaranteed to give the same results as the original method, and it empirically often requires much fewer steps for convergence, a property we empirically investigate in this section. We first show that our method achieves consistent more than 4x speedup when applied to a variety of 3D representations and different score distillation frameworks. We then conduct ablation studies on the proposed practical decisions.

\subsection{Accelerating Text-to-3D Generation}

We choose baselines that represent the most prominent 3D representations, namely NeRF from DreamFusion~\citep{Poole2022-sm}, DMTet from Magic3D~\citep{Lin2022-ha}, SDF from coarse-stage TextMesh~\citep{Tsalicoglou2023-js} (see \appref{app:exp} for detail), and 3D Gaussian Splatting from DreamGaussian~\citep{Tang_undated-od}, and directly apply our wrapper to them. In addition, we show that our algorithm can similarly be adapted to the recently proposed VSD from ProlificDreamer~\citep{Wang2023-lp}. Our \model{}  wrapper is modular and agnostic to the calculation of gradient updates via score distillation, and we test its performance by comparing each of the baselines' runtime and quality with and without our wrapper.

\mypara{Data and metrics.} We use 30 prompts from the DreamFusion gallery to test each algorithm. We run each framework with each individual prompt and record its wallclock runtime in seconds. Following \citep{Tsalicoglou2023-js,Jain2022-hr,Mohammad_Khalid2022-ew}, we use CLIP R-Precision~\citep{Park2021-zj} for measuring semantic alignment between the generated asset and its input prompt. CLIP R-Precision is the retrieval accuracy of a prompt from the set of prompts given a generated image conditioned on this prompt, and we use top-1 retrieval accuracy for all experiments. We additionally use CLIP FID score~\citep{Kynkaanniemi2022-yx} compared against the ImageNet 2012 validation set~\citep{Russakovsky2014-sq} for generation quality. Speedup is calculated as the ratio of wall-clock runtime for baseline to our method -- higher is better.

\mypara{Evaluation.} As larger batch size can lead to higher-quality generation~\citep{Lin2022-ha,Chen2023-sb,Pan2023-jq, Wang2023-lp}, we use batch size 16 for all baselines but ProlificDreamer, for which, due to memory constraints, we use batch size 8 for the first 5000 steps (rendered at $64\times 64$ resolution) and batch size 2 for the remaining steps (rendered at $512\times 512$ resolution). We use 8 NVIDIA A100 PCIe GPUs for all experiments, which default to window size 7. More details can be found in \appref{app:exp}. 
We evaluate R-Precision and $\text{FID}_{\text{CLIP}}$ using all rendered images from all generated shapes. Quantitative results can be found in \tabref{tab:text-to-3d} and qualitative results are shown in \figref{fig:text-to-3d}.

\begin{table}[t]

\centering

\resizebox{\linewidth}{!}{
\begin{tabular}{lcccc}
\toprule
 &  R-Precision $\uparrow$ &  FID $\downarrow$  & Runtime (s) $\downarrow$ & Speedup\\
\cmidrule{2-5}
DreamFusion~\citep{Poole2022-sm}   & \textbf{82.70}  & 60.91  & 8080  & \multirow{ 2}{*}{4.22x}\\
\textbf{+ \model} &  79.57 & \textbf{ 59.88} & \textbf{1910 }&\\
\midrule
Magic3D~\citep{Lin2022-ha}   & \textbf{88.45} &  59.24 & 13470 & \multirow{ 2}{*}{4.17x}\\
\textbf{+ \model} & 87.18  & \textbf{59.19} &\textbf{ 3230} &  \\
\midrule
TextMesh~\citep{Tsalicoglou2023-js}   & \textbf{88.06}  & 64.19 & 7690 & \multirow{ 2}{*}{ 4.18x}\\
\textbf{+ \model} & 84.43   & \textbf{62.12} & \textbf{1830} & \\
\midrule
DreamGaussian~\citep{Tang_undated-od}  & 83.28 & 58.33 & 700 & \multirow{ 2}{*}{ 4.67x} \\
\textbf{+ \model} & \textbf{83.85} &  \textbf{58.26}  & \textbf{150} &\\
\midrule
ProlificDreamer~\citep{Wang2023-lp}   & 94.11  & 49.66 & 7390 & \multirow{ 2}{*}{4.69x}\\
\textbf{+ \model} & \textbf{96.12} & \textbf{49.65} & \textbf{3710} &\\
\midrule
\bottomrule
\end{tabular}
}
\caption{Quantitative evaluation on 30 prompts from the DreamFusion gallery. Runtime is reported in seconds. Our method achieves competitive quality while provide more than 4x speedup.}
\label{tab:text-to-3d}

\end{table}

\begin{figure*}[t]
    \centering
    \begin{subfigure}[b]{0.33\textwidth}
        \centering
        \includegraphics[width=\linewidth,trim=1 1 1 1,clip]{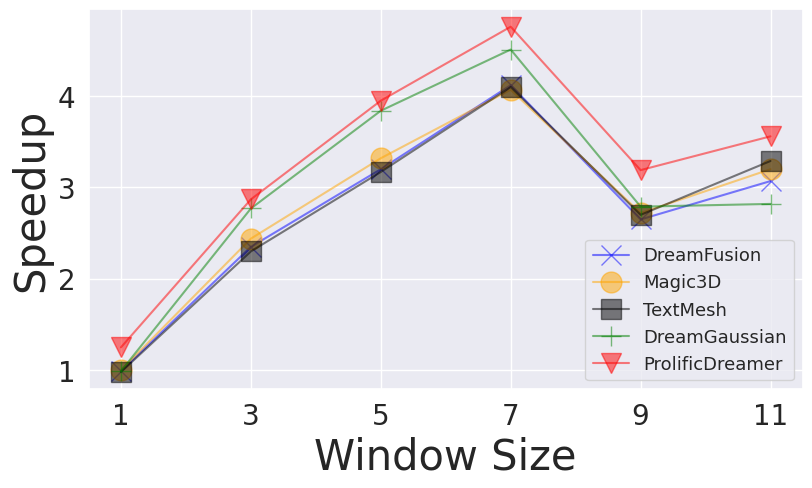}
        \caption{Ablation on window size.}
        \label{fig:ablation_gpu}
    \end{subfigure}%
    \begin{subfigure}[b]{0.33\textwidth}
        \centering
        \includegraphics[width=\linewidth,trim=1 1 1 1,clip]{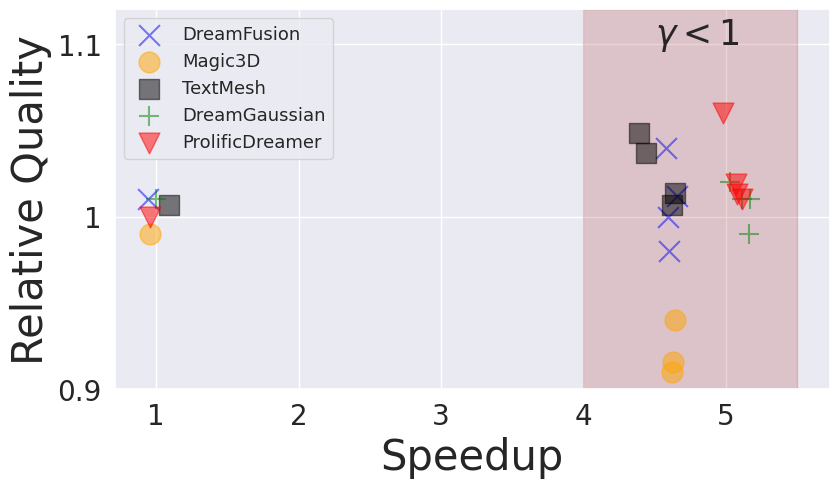}
        \caption{Ablation on threshold EMA rate.}
        \label{fig:ablation_ema}
    \end{subfigure}%
    \begin{subfigure}[b]{0.33\textwidth}
        \centering
        \includegraphics[width=\linewidth,trim=1 1 1 1,clip]{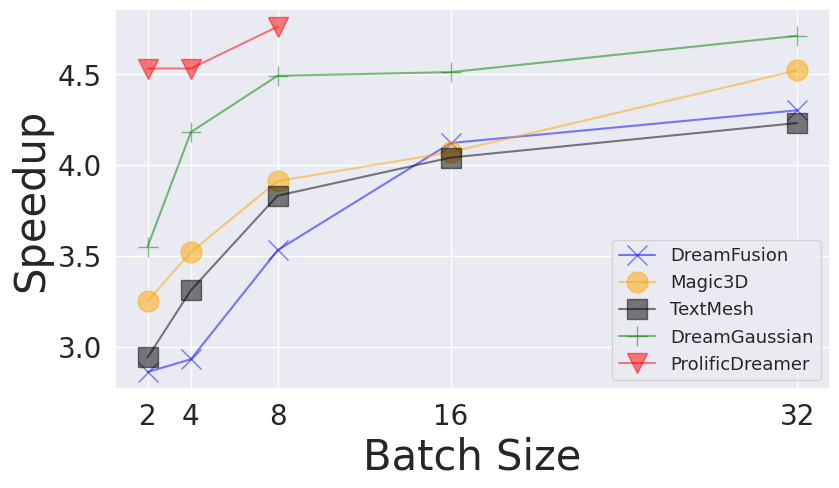}
        \caption{Ablation on batch size.}
        \label{fig:ablation_bs}
    \end{subfigure}%
    \caption{Ablation studies on practical choices. Speedup is the ratio of baseline wall-clock runtime to our wall-clock runtime. Relative quality is the ratio of baseline $\text{FID}_\text{CLIP}$ to our $\text{FID}_\text{CLIP}$.}
    \label{fig:ablation}
\end{figure*}

Notice that with competitive generation quality, we consistently achieve more than 4x speedup for all frameworks and algorithms, with the most speedup for ProlificDreamer. This is in line with the intuition that our parallelization algorithm is more effective when each iteration's GPU workload becomes heavier, which is the case for ProlificDreamer due to it keeping and updating another LoRA model on the fly. The heavy workload is effectively delegated to different GPUs. In theory, the heavier the GPU workload is, the more the optimization benefits from our framework. Also note that DreamGaussian experiences competitive speedup as ProlificDreamer. This is because its original implementation requires sequential rendering of each instance within a batch. This scales its GPU runtime linearly with batch size, so our framework gives better speedup. We also notice that our results do not exactly match those of baselines. This is likely due to the fixed-point error not being low enough for converging to the fixed-point solution, and Adam takes the parameters to slightly different but equally valid local optima through momentum-based gradient updates.

\if\arxiv1

\subsection{Accelerating Image-to-3D Generation}\label{app:image-to-3d}

Many works have also explored score distillation for Image-to-3D generation using 2D-diffusion finetuned on view-dependent data~\citep{Liu2023-kk,Qian2023-rv,Liu2023-ut}. Among the most popular approaches is Zero-1-to-3~\citep{Liu2023-kk}, which finetunes a large-scale diffusion model for novel-view synthesis given a single image and novel-view embeddings. It also serves as a powerful 3D-aware prior for score distillation, which luckily our framework can be directly applied to. In this section, we investigate our framework's application to the Image-to-3D generation task.

For evaluation, we choose NeRF~\citep{Mildenhall2021-wv} and 3D Gaussian Splatting~\citep{Kerbl2023-kn} as the two representative examples for 3D representations with constant and changing dimensions during optimization. Each representation is equipped with Zero-1-to-3 and a source image for generating a novel 3D object, and we show that our framework can achieve substantial speedup when applied to Zero-1-to-3 while retaining generation quality. For both representations, we use batch size 16 unless otherwise noted, and we run SDS for 1200 steps and 500 steps respectively (details in \appref{app:exp}). We show 3 examples for each representation in~\figref{fig:image-to-3d}, where we compare novel views of the 3D results from Zero-1-to-3 and the results from Zero-1-to-3 accelerated by \model. 

We observe that our framework can achieve almost identical generation output with much shorter runtime for both representations. The wallclock time speedup is consistently more than 3x and 4x the original runtime. NeRF achieves lower speedup than the Text-to-3D counterpart due to the lower number of total steps compared to Text-to-3D generation (\eg 25,000 steps), so the time for the costly initial model and data preloading for all GPUs is not effectively amortized. 3D Gaussian Splatting is less affected thanks to its lightweight representation conducive to fast cross-GPU communication and  DreamGaussian's~\citep{Tang_undated-od} simple implementation with minimal initial allocation cost.

\begin{figure*}[t]
    \centering
    \setlength{\tabcolsep}{0mm}
     \begin{tabular}{@{}>{\kern-\tabcolsep}c@{}c@{}c@{}c@{}c@{}c<{\kern-\tabcolsep}@{}}
       Source & NeRF~\citep{Mildenhall2021-wv}  & + \model & Source & 
 3DGS~\citep{Kerbl2023-kn} & + \model \\
         \begin{minipage}{.14\textwidth}
          \includegraphics[width=\linewidth]{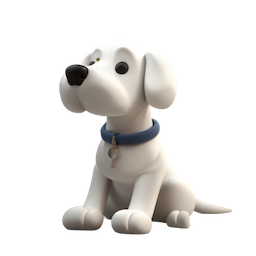} 
        \end{minipage} & 
        \begin{minipage}{.14\textwidth}
          \includegraphics[width=\linewidth]{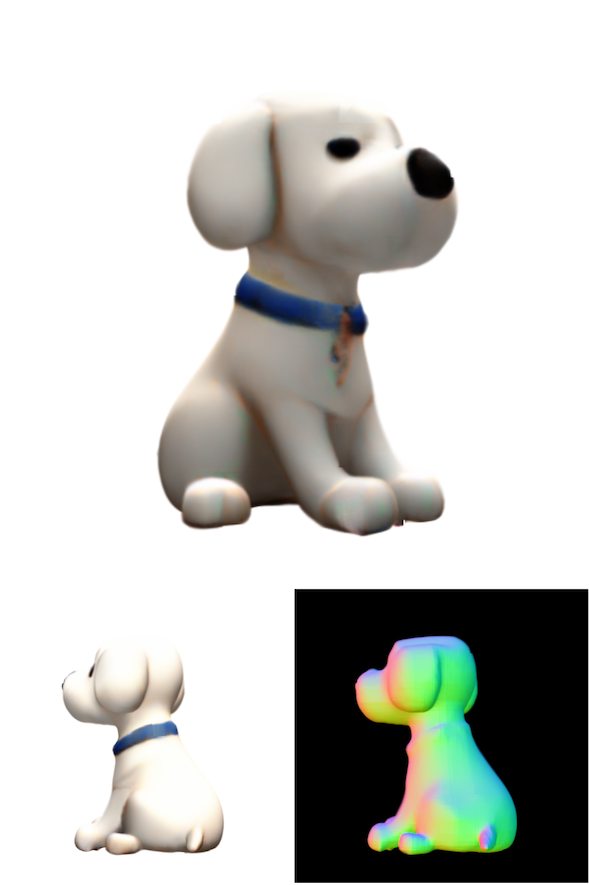} 
        \end{minipage} & 
        \begin{minipage}{.14\textwidth}
          \includegraphics[width=\linewidth]{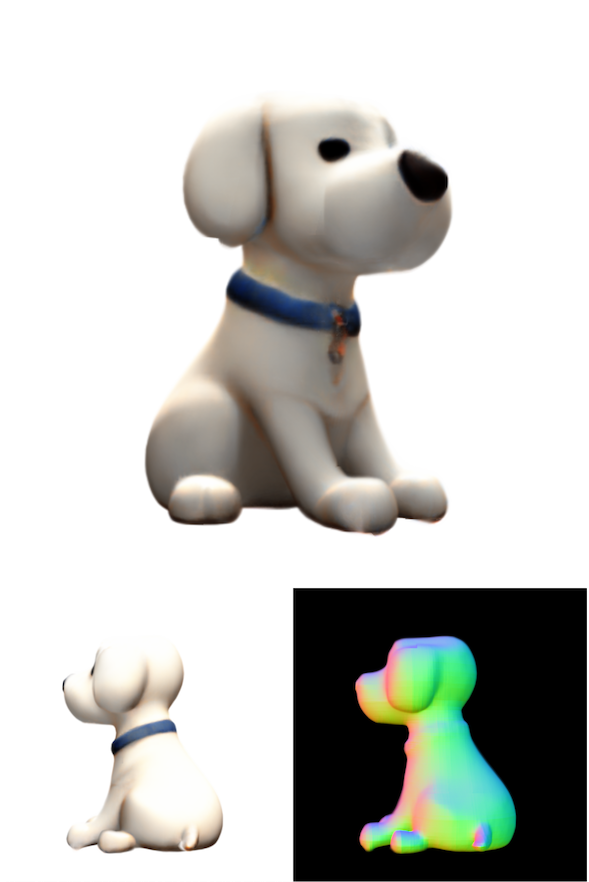}\end{minipage}  &
          \hfill
         \begin{minipage}{.14\textwidth}
          \includegraphics[width=\linewidth]{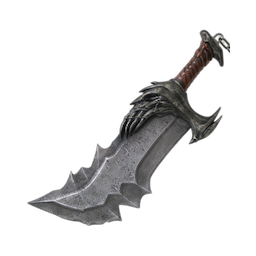} 
        \end{minipage} & 
        \begin{minipage}{.14\textwidth}
          \includegraphics[width=\linewidth]{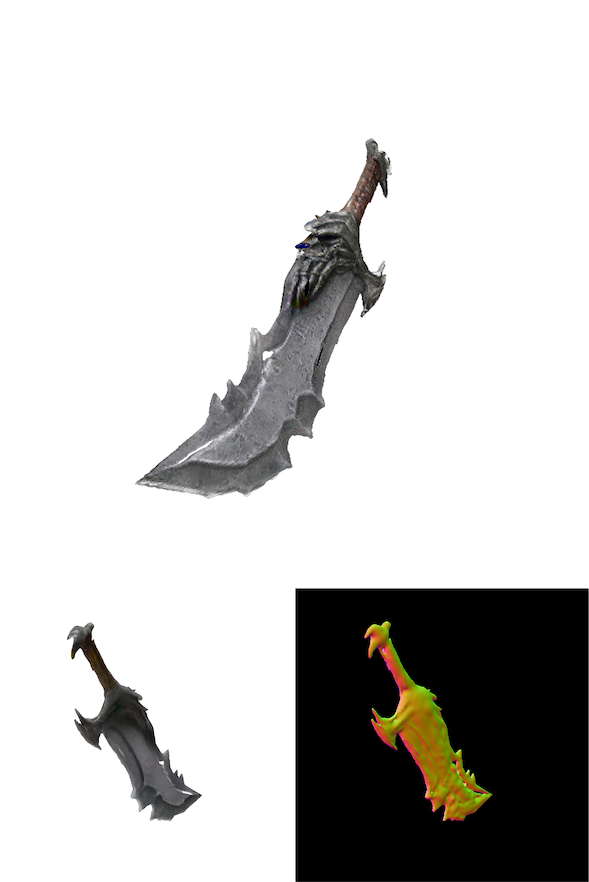} 
        \end{minipage} & 
        \begin{minipage}{.14\textwidth}
          \includegraphics[width=\linewidth]{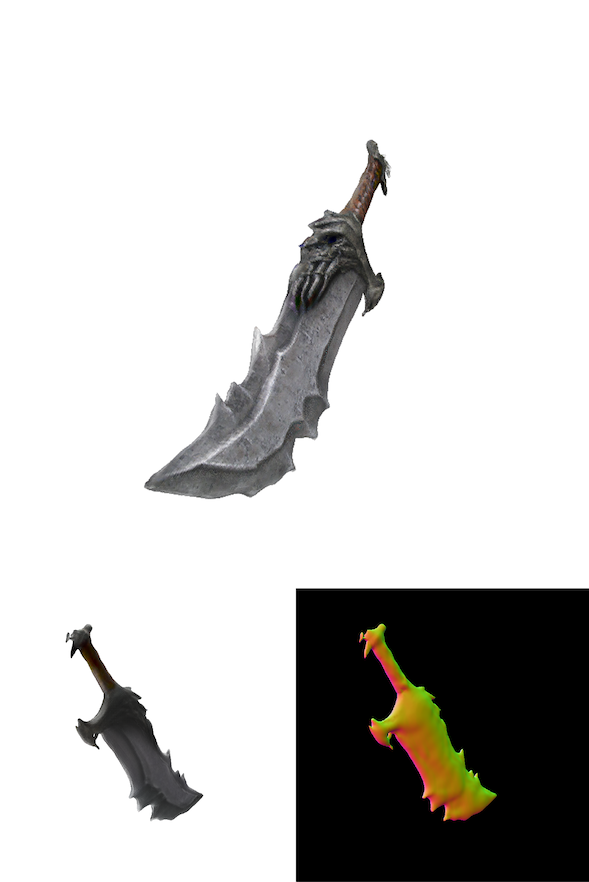}\end{minipage} \\
        &  \cellcolor{Gainsboro!60} 11 min 38 sec & \cellcolor{Gainsboro!60}\textbf{3 min 45 sec} & & \cellcolor{Gainsboro!60} 6 min 1 sec & \cellcolor{Gainsboro!60}\textbf{1 min 13 sec } \\
        \begin{minipage}{.14\textwidth}
          \includegraphics[width=\linewidth]{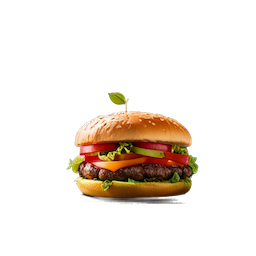} 
        \end{minipage} & 
        \begin{minipage}{.14\textwidth}
          \includegraphics[width=\linewidth]{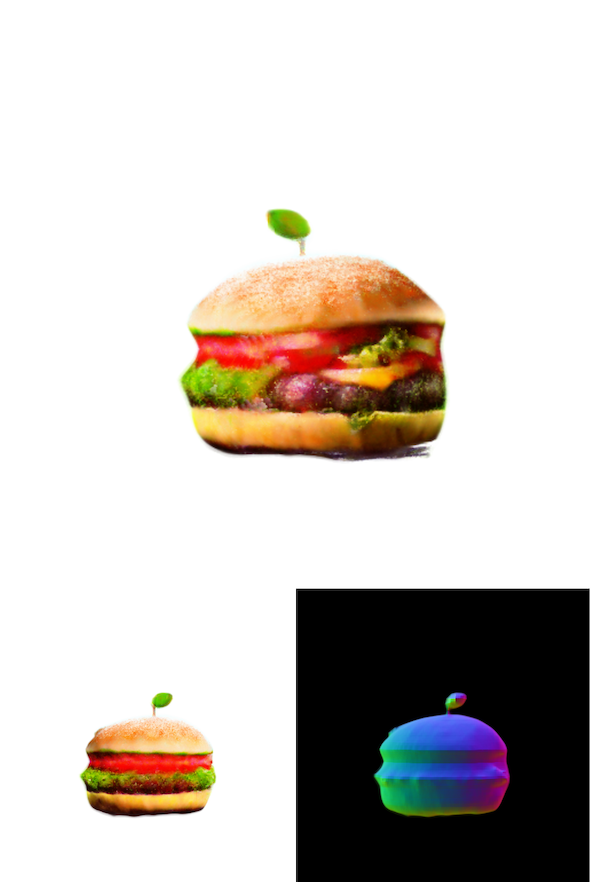} 
        \end{minipage} & 
        \begin{minipage}{.14\textwidth}
          \includegraphics[width=\linewidth]{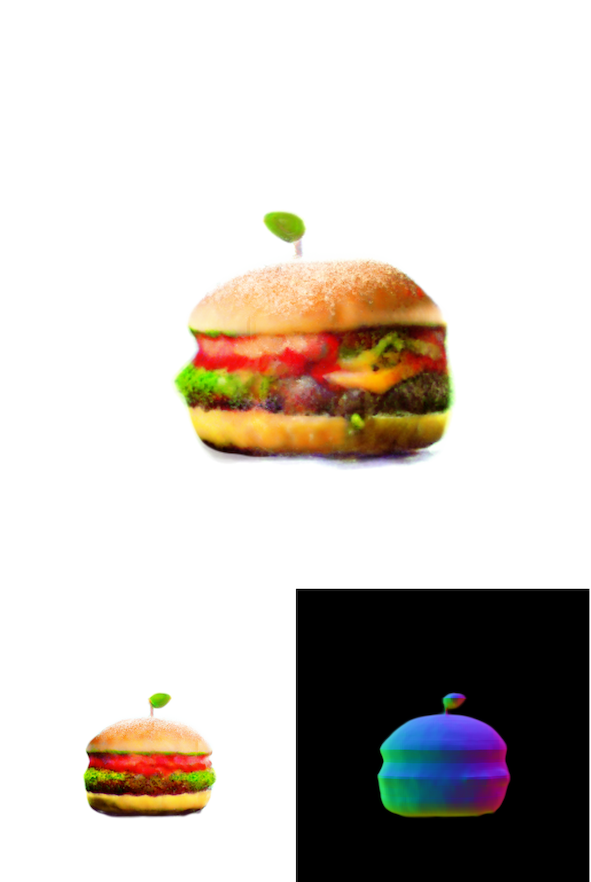}\end{minipage}  &
          \hfill
         \begin{minipage}{.14\textwidth}
          \includegraphics[width=\linewidth]{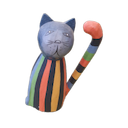} 
        \end{minipage} & 
        \begin{minipage}{.14\textwidth}
          \includegraphics[width=\linewidth]{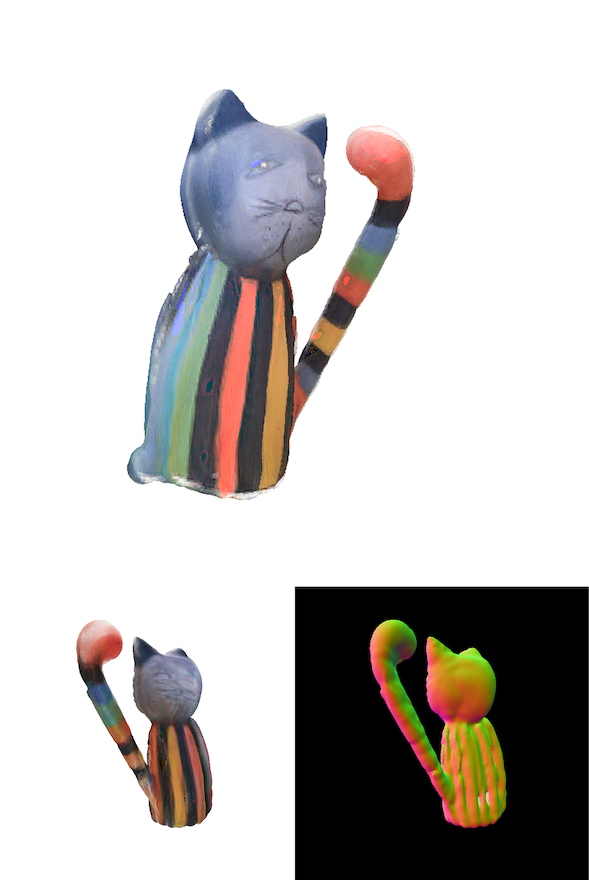} 
        \end{minipage} & 
        \begin{minipage}{.14\textwidth}
          \includegraphics[width=\linewidth]{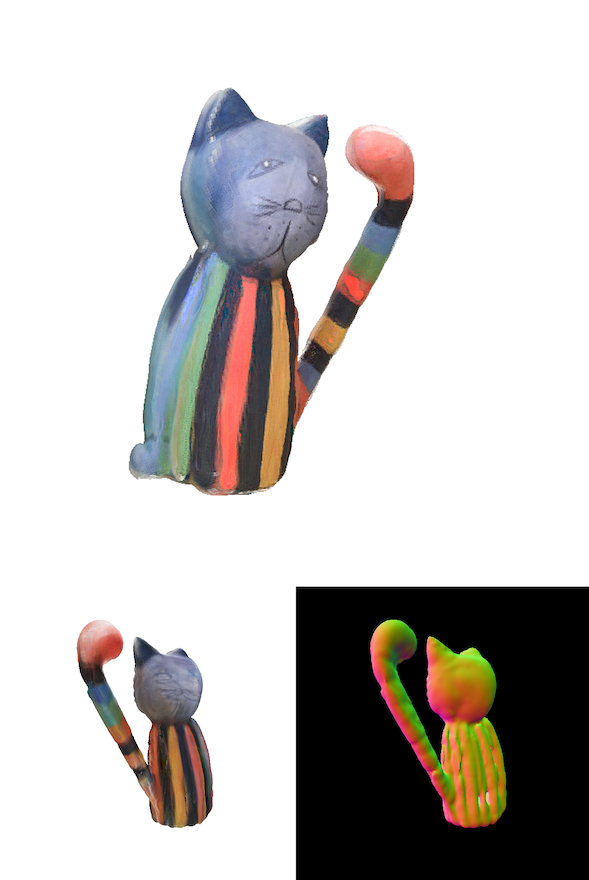}\end{minipage} \\
        &  \cellcolor{Gainsboro!60} 10 min 53 sec & \cellcolor{Gainsboro!60}\textbf{3 min 17 sec} & & \cellcolor{Gainsboro!60} 5 min 56 sec & \cellcolor{Gainsboro!60}\textbf{1 min 15 sec} \\
        \begin{minipage}{.14\textwidth}
          \includegraphics[width=\linewidth]{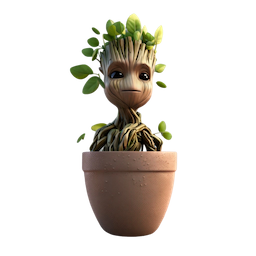} 
        \end{minipage} & 
        \begin{minipage}{.14\textwidth}
          \includegraphics[width=\linewidth]{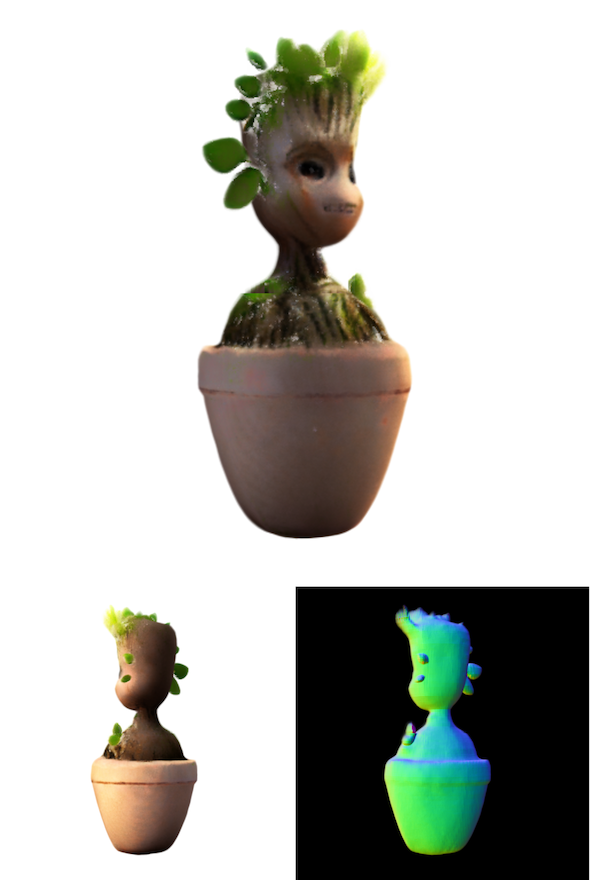} 
        \end{minipage} & 
        \begin{minipage}{.14\textwidth}
          \includegraphics[width=\linewidth]{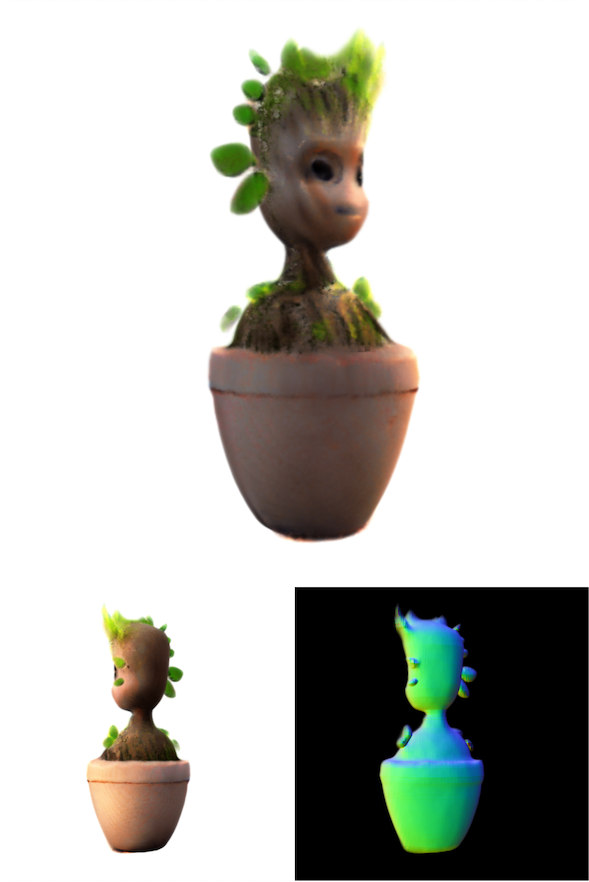}\end{minipage}  &
          \hfill
         \begin{minipage}{.14\textwidth}
          \includegraphics[width=\linewidth]{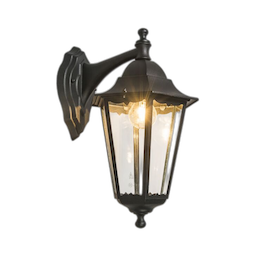} 
        \end{minipage} & 
        \begin{minipage}{.14\textwidth}
          \includegraphics[width=\linewidth]{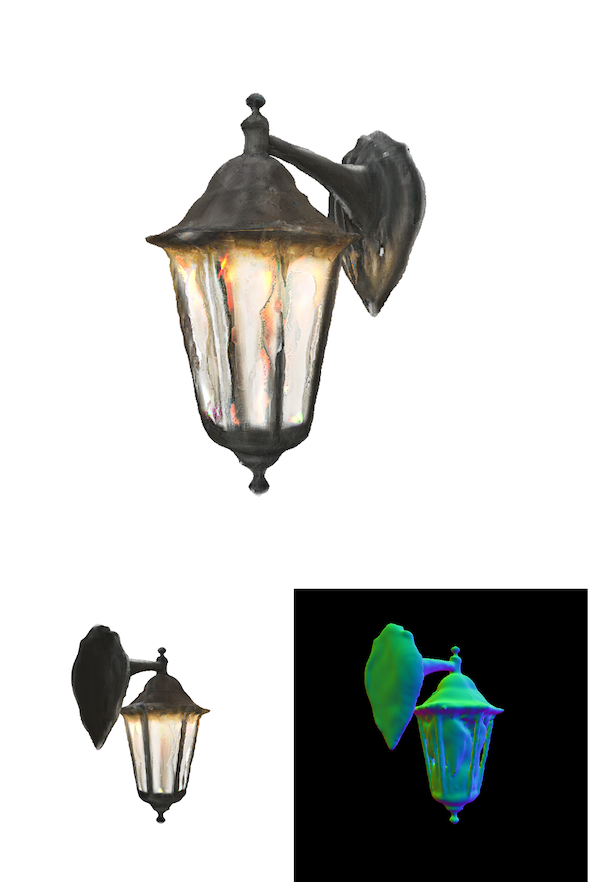} 
        \end{minipage} & 
        \begin{minipage}{.14\textwidth}
          \includegraphics[width=\linewidth]{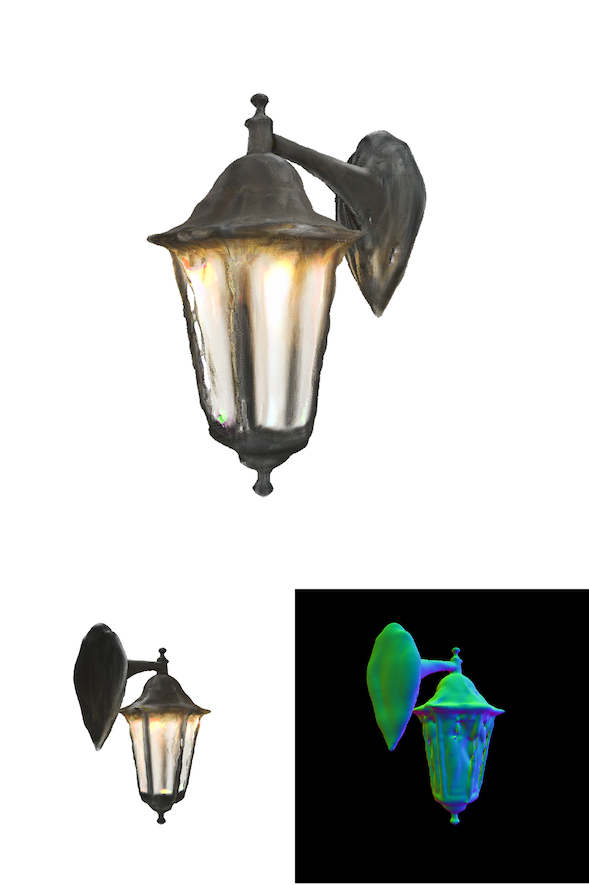}\end{minipage} \\
        &  \cellcolor{Gainsboro!60} 10 min 51 sec & \cellcolor{Gainsboro!60}\textbf{3 min 11 sec} & & \cellcolor{Gainsboro!60} 5 min 58 sec & \cellcolor{Gainsboro!60}\textbf{1 min 15 sec}\\
          \rowcolor{Gainsboro!60}
      Avg. Speedup  &  \multicolumn{2}{c}{3.28x} &  Avg. Speedup & \multicolumn{2}{c}{4.8x}   \\
         
     \end{tabular}
    \caption{Visual comparisons for Image-to-3D generation using Zero-1-to-3~\citep{Liu2023-kk}. Our method achieves more than 3x and 4x average speedup respectively when applied to Zero-1-to-3 while achieving almost identical generation results.}
    \label{fig:image-to-3d}
    \vspace{-4mm}
\end{figure*}

\fi
\subsection{Ablation Studies}
We further conduct ablation studies for the practical choices to investigate their effectiveness.

\mypara{Effect of window size.} By default, we set the window size to be 1 less than the number of GPUs, since we need the remaining one for sequential rollout. Allowing all other GPUs to do independent work intuitively maximizes speedup. However, how does the speedup scale when window size is independent of the number of GPUs? For our experiment in \figref{fig:ablation_gpu}, we assume access to 8 NVIDIA A100 PCIe GPUs and adjust the window size from 1 to beyond 7. The speedup of our framework peaks at window size 7, because we fully utilize all GPU resources for maximum parallelization. We also observe a drop in speedup for window size beyond 7. We hypothesize that this is due to longer sliding windows not advancing far enough under GPU resource constraints while requiring more FLOPs, thus causing slower speedup.

\mypara{Effect of threshold adaptivity.} The adaptivity parameter aims to stabilize speedup while maintaining generation quality. We show in \figref{fig:ablation_ema} its ablation study. Relative quality is calculated as the ratio of baseline $\text{FID}_\text{CLIP}$ to our $\text{FID}_\text{CLIP}$ -- a higher ratio means better relative quality. For each model, 5 settings are tested with $\gamma\in \{0.2,0.4,0.6,0.8,1.0\}$ and initial thresholds set to 1\% of the default thresholds. This is to maximally bottleneck our framework to test how much $\gamma$ counteracts an ill-initialized threshold. We observe a large gap between results from $\gamma=1$ and $\gamma<1$ (in red region), implying as long as $\gamma<1$ (\ie our framework is adaptive), all models achieve similar speedup without significantly degrading quality.


\mypara{Effect of batch size.} As SDS/VSD requires evaluating expectation over diffusion time $t$, one is motivated to use a large batch size of $t$ for more accurate estimate of the guidance score, which can lead to higher-quality generation~\citep{Lin2022-ha,Chen2023-sb,Pan2023-jq, Wang2023-lp}. As such, this increases computational demand for GPU per iteration, a regime where our method is particularly beneficial. As shown \figref{fig:ablation_bs}, the speedup of our method scales with increasing batch sizes because of the increase in computational intensity per iteration. In particular, our method benefits VSD more even when batch size is small because the LoRA training is costly and is effectively amortized across GPUs because their parameters are never communicated across processes. 

\section{Conclusion}


In this work, we introduce \model, a drop-in acceleration framework for all existing score distillation-based text-to-3D generation methods. \model{} can achieve more than 4x speedup for various predominant 3D representations and benefits more from heavier demand for GPU computation per iteration. We hope our framework serves as a significant step towards usable high-quality text-to-3D methods and inspires more advancements to come.
\if\arxiv1
\section{Contribution and Acknowledgement}
This research is done at Pika Labs and supported in part by NSF(\#1651565), ARO (W911NF-21-1-0125), ONR (N00014-23-1-2159), CZ Biohub, HAI. Linqi Zhou led and worked on all parts of the project and Andy Shih proposed the initial generalized Picard iteration and contributed to framework formulation and initial experiments. We thank additional help and discussion with Chenlin Meng, Stefano Ermon, Felix Petersen, Wanqiao Xu and other colleagues at Stanford University.
\fi

{
    \small
    \bibliographystyle{ieeenat_fullname}
    \bibliography{egbib}
}

\clearpage
\setcounter{page}{1}
\maketitlesupplementary

\appendix
\section{Theoretical Results}\label{app:proof}

We show that the iteration rule in~\cref{eq:general-rollout} satisfies the fixed-point property, and is guaranteed to converge to the true solution, i.e., the solution obtained by sequential computation which we denote as $\theta^\star_{0}, \ldots, \theta^\star_{T}$. Our result follows from a simple observation via induction. First, assume that after $k$ fixed-point iterations, $\theta^k_{\tau} = \theta^\star_{\tau}$ for all $\tau \leq k$. We always initialize $\theta^k_{0}$ to be $\theta^\star_{0}$, so our inductive hypothesis trivially holds for $k=0$. Next we examine the iteration rule for iteration $k+1$ and simplify terms using the pseudo-inverse property that $h^\dagger(s(\theta^\star_{\tau}), \theta^\star_\tau) = g(\theta^\star_\tau) = \theta^\star_{\tau+1}$.

\begin{align*}\label{eq:general-rollout-induction}
    \theta_{\tau+1}^{k+1} &= h^\dagger\left(s(\theta_{\tau}^{k}), \dots h^\dagger\left(s(\theta_{1}^{k}),  h^\dagger(s(\theta_{0}^{k}), \theta_0^{k})\right) \right) \\
    \theta_{\tau+1}^{k+1} &= h^\dagger\left(s(\theta_{\tau}^{\star}), \dots h^\dagger\left(s(\theta_{1}^{\star}),  h^\dagger(s(\theta_{0}^{\star}), \theta_0^{\star})\right) \right) \\
    \theta_{\tau+1}^{k+1} &= h^\dagger\left(s(\theta_{\tau}^{\star}), \dots h^\dagger\left(s(\theta_{1}^{\star}),  \theta_{1}^{\star} \right) \right) \\
    \theta_{\tau+1}^{k+1} &= h^\dagger\left(s(\theta_{\tau}^{\star}), \dots \theta_{2}^{\star} \right) \\
    \theta_{\tau+1}^{k+1} &= h^\dagger\left(s(\theta_{\tau}^{\star}), \theta_{\tau}^{\star} \right) \\
    \theta_{\tau+1}^{k+1} &= \theta_{\tau+1}^{\star}
\end{align*}

The above derivation shows that the inductive hypothesis extends to all $\tau \leq k+1$. Therefore, after at most $T$ iterations the fixed-point iteration will converge to the true solution. In practice, however, the fixed-point iteration may converge with much fewer number of iterations.

\section{Algorithm}\label{app:algo}

\begin{algorithm}[t]
\caption{\model}\label{alg:algo}
\begin{algorithmic}
\State {\bf Input:} 3D model with initial parameter $\theta$, parameter dimension $D$, total time $T$, initial threshold $e$, error aggregation function $M$, Adam optimizer, EMA update $\gamma$, number of GPUs $n$.
\State {\bf Output:} Score distillation output
\State $\tau,k,p \leftarrow 0,0, n-1$
\State $\theta_i^0 \leftarrow \theta \quad \forall i\in [0, p-1]$ 
\State Initialize diffusion guidance (and optionally, LoRA for VSD) on $p$ GPUs.
\While{$\tau < T$}
\State Dispatch $\{\theta_{\tau+j}^k\}_{j=0}^{p-1}$ to different GPUs and gather $\{s(\theta_{\tau+j}^k)\}_{j=0}^{p-1}$ with seed $\tau$ (Optionally, update LoRA separately on each GPU). {\color{Purple} // Gather $\gradnd{\mathcal{L}_{\text{SDS}}}{\theta}$ or $\gradnd{\mathcal{L}_{\text{VSD}}}{\theta}$ }
\State $\theta_{\tau+j+1}^{k+1}\leftarrow h^\dagger(s(\theta_{\tau+j}^k),  \dots h^\dagger(s(\theta_{\tau}^k), \theta_{\tau}^k) \dots ), \quad \forall j\in [0, p-1]$ {\color{Purple}  } 
\State error $\leftarrow \{\frac{1}{D}d(\theta_{\tau+j}^{k+1}, \theta_{\tau+j}^k)^2\}_{j=1}^p$
\State skip $\leftarrow \min(\{j: \frac{1}{D}d(\theta_{\tau+j}^{k+1}, \theta_{\tau+j}^k)^2 > e, \quad \forall j\in [1,p]\}\cup \{p\} )$ 
\State $\theta_{\tau+j}^{k+1} \leftarrow \theta_{\tau+p}^{k+1} \quad \forall j\in [p, \text{skip}+p]$  {\color{Purple} // New window }
\State $\tau \leftarrow \tau + \text{skip},\quad k\leftarrow  k+1$
\State $e \leftarrow \gamma e + (1 - \gamma) * M(\text{error})$ {\color{Purple}// Adaptive threshold}
\State $p\leftarrow \min(p, T-\tau)$
\EndWhile

\State \Return $\theta_{T}^k$
\end{algorithmic}
\end{algorithm}

A detailed algorithm is presented in \algoref{alg:algo}. We emphasize that since the computational units $s(\cdot)$ do not have nested dependencies, they can be computed \textit{in parallel}. On the other hand, the $h^\dagger(\cdot)$ must be unrolled sequentially. Therefore, for speedup with parallel computation, $s(\cdot)$ should be chosen to contain most of the computational cost.

\section{Additional Details on Practical Decisions}\label{app:practice}

\mypara{Sliding window.} The sliding window scheme can be better understood with a simple example. For a window of size 3 containing $\{\theta_\tau^k, \theta_{\tau+1}^k, \theta_{\tau+2}^k\}$ with corresponding drifts $\{s(\theta_\tau^k), s(\theta_{\tau+1}^k), s(\theta_{\tau+2}^k)\}$, one slides the window forward by one if $\norm{\theta_{\tau+1}^{k+1}- \theta_{\tau+1}^k}^2 \ge e$ and slides the window by two if $\norm{\theta_{\tau+1}^{k+1}- \theta_{\tau+1}^k}^2 < e$ and $\norm{\theta_{\tau+2}^{k+1}- \theta_{\tau+2}^k}^2 \ge e$. 

We employ $\gL_2$ norm on 3D parameter space in practice. In the case of generalized iteration, $\theta_{\tau}^{k+1}$ and $\theta_{\tau}^{k}$ may lie in different dimensions. We then calculate the  $\gL_2$ error between the two by $\norm{\theta_{\tau+1}^{k+1}- \text{unproj}(\theta_{\tau+1}^k)}^2$. We denote this (squared) distance as $d(\theta_{\tau+1}^{k+1}, \theta_{\tau+1}^k)^2$ hereon. For the case of Adam optimizer, we choose to only measure the distance between $\theta$ parameters and ignore momentum parameters for convergence checking.

\mypara{Eliminating stochasticity.} We simply set the seed for time step $\tau$ to be $\tau$. For increased variability, we can sample a random seed $s$ before running the algorithm and set the seed for time step $\tau$ to be $\tau+s$. 

\mypara{Parallelizing Variational Score Distillation.} We reuse all settings from \citet{Wang2023-lp} for each independent copies. 
\begin{table*}[th]
    \centering
    
\resizebox{\linewidth}{!}{
    \begin{tabular}{cccccc}
    \toprule
         & DreamFusion\citep{Poole2022-sm} & Magic3D \citep{Lin2022-ha} & TextMesh \citep{Tsalicoglou2023-js}& DreamGaussian \citep{Tang_undated-od} & ProlificDreamer \citep{Wang2023-lp} \\
         \midrule
       Guidance Model &  DeepFloyd & DeepFloyd+SD-2.1 & DeepFloyd & SD-2.1 & SD-2.1 \\
       Window Size $p$ & 7 & 7 & 7& 7& 7\\
       Initial Threshold $e$  $(\times 10^{-6})$ & 5 & 5& 5 & 500& 50   \\
      Adaptivity $\gamma$& 0.9& 0.9& 0.9& 0.9& 0.9\\
      Error Aggregation Function $M$ & median & median & median & median & mean\\
      Batch Size& 16 & 16 & 16 &16 & [8,2],8,1\\
       \bottomrule
    \end{tabular}
    }
    \caption{Default parameters used for Text-to-3D generation.}
    \label{tab:parameters}
\end{table*}

\section{Experiments}\label{app:exp}
Our code is based on \Verb"threestudio"~\citep{Liu_undated-fp}\footnote{\url{https://github.com/threestudio-project/threestudio}} and for all the baselines we reuse their settings. Note that for TextMesh, \Verb"threestudio" only implements the coarse-stage generation. Therefore, as a wrapper around the package, we only experiment with coarse-stage TextMesh to demonstrate our technique's effectiveness on this representation. For ProlificDreamer, for quantitative metrics, due to time limitations, we only report the runtime and quality metrics of the first stage (using NeRF). We observe similar speedup for its second/third stage during refinement. The qualitative comparisons, however, include the second and third stage refinement. The second stage also has batch size of 8 but for the third stage, we use batch size of 1 (since anything larger is prohibitively slow for baselines). Similarly, we build a lightweight wrapper for 3D Gaussian Splatting using the DreamGaussian implementation\footnote{\url{https://github.com/dreamgaussian/dreamgaussian}}. 

We summarize the default parameter settings relevant for our algorithm in \tabref{tab:parameters} for Text-to-3D generation. All other parameters are held fixed from \Verb"threestudio" and \Verb"DreamGaussian" implementations. Note that for DreamFusion and Magic3D, we use DeepFloyd for coarse-stage generation and we use SD-2.1 for refinement.

For Image-to-3D generation, we also use the Zero-1-to-3 implementation from \Verb"threestudio", which uses NeRF as the 3D representation, and \Verb"DreamGaussian", which uses 3D Gaussian Splatting, respectively. For NeRF, we reuse settings from \Verb"threestudio". For SDS, we progressively increase the rendering size in the order of \{64, 128, 256\} and use batch size \{16, 16, 10\} at step \{0, 600, 900\}. The single image size is also changed in the order of \{128, 256, 512\}. For 3D Gaussian Splatting, we use batch size 16 and reuse other settings from \Verb"DreamGaussian" implementation. We summarize relevant parameter settings in \tabref{tab:parameters-im}. 

\begin{table}[h]
    \centering
    
\resizebox{\linewidth}{!}{
    \begin{tabular}{ccc}
    \toprule
         & NeRF~\citep{Mildenhall2021-wv} & 3DGS~\citep{Kerbl2023-kn}  \\
         \midrule
       Guidance Model &  Zero-1-to-3 & Zero-1-to-3 \\
       Window Size $p$ & 7 & 7 \\
       Initial Threshold $e$  $(\times 10^{-6})$ & 30 & 500   \\
      Adaptivity $\gamma$& 0.9& 0.9 \\
      Error Aggregation Function $M$ & median & median \\
      Batch Size& 16,16,10 & 16 \\
       \bottomrule
    \end{tabular}
    }
    \caption{Default parameters used for Image-to-3D generation using Zero-1-to-3~\citep{Liu2023-kk}.}
    \label{tab:parameters-im}
\end{table}

\mypara{Evaluation settings.} For Text-to-3D, we render a learned 3D shape from 12-degree elevation angle and 120 evenly-spaced azimuth angles all around. Each image is paired with the shape's input prompt. We use CLIP-B/32 for both R-Precision and FID calculation. The reference statistics for FID is the ImageNet 2012 validation set.

\if\arxiv0
\newpage

\begin{figure*}[t]
    \centering
    \setlength{\tabcolsep}{0mm}
     \begin{tabular}{@{}>{\kern-\tabcolsep}c@{}c@{}c@{}c@{}c@{}c<{\kern-\tabcolsep}@{}}
       Source & NeRF~\citep{Mildenhall2021-wv}  & + \model & Source & 
 3DGS~\citep{Kerbl2023-kn} & + \model \\
         \begin{minipage}{.14\textwidth}
          \includegraphics[width=\linewidth]{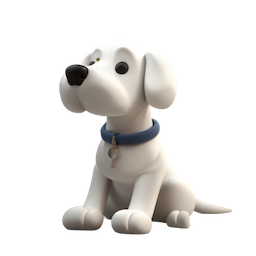} 
        \end{minipage} & 
        \begin{minipage}{.14\textwidth}
          \includegraphics[width=\linewidth]{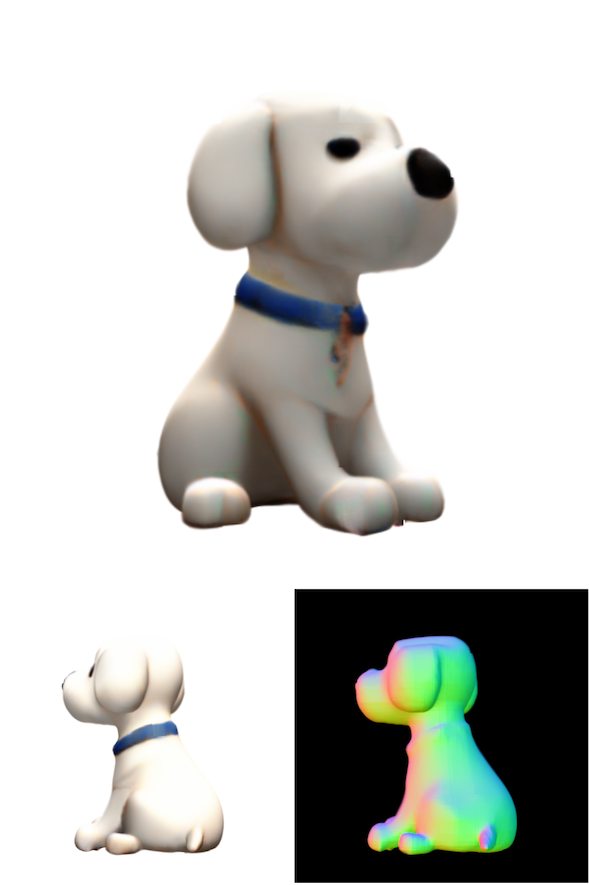} 
        \end{minipage} & 
        \begin{minipage}{.14\textwidth}
          \includegraphics[width=\linewidth]{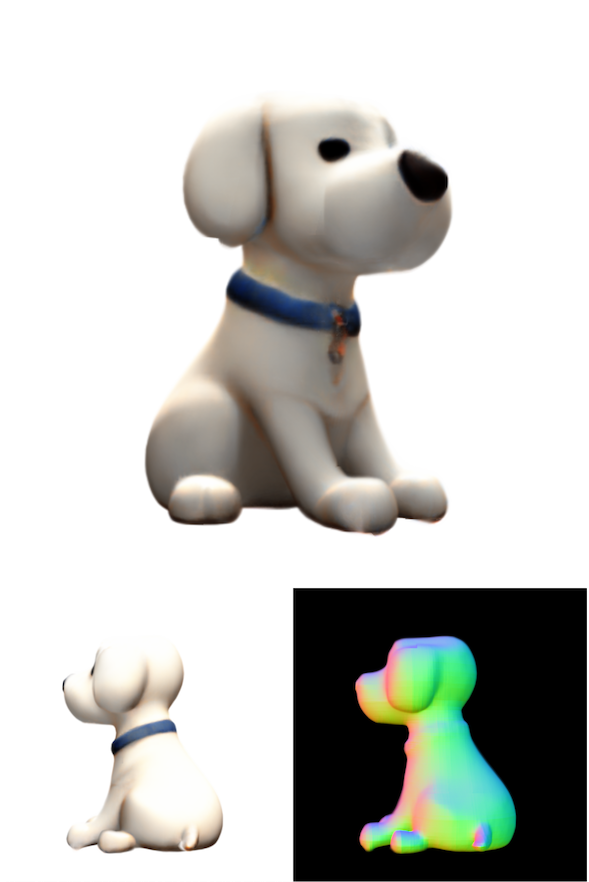}\end{minipage}  &
          \hfill
         \begin{minipage}{.14\textwidth}
          \includegraphics[width=\linewidth]{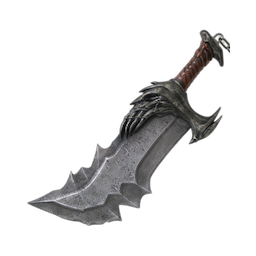} 
        \end{minipage} & 
        \begin{minipage}{.14\textwidth}
          \includegraphics[width=\linewidth]{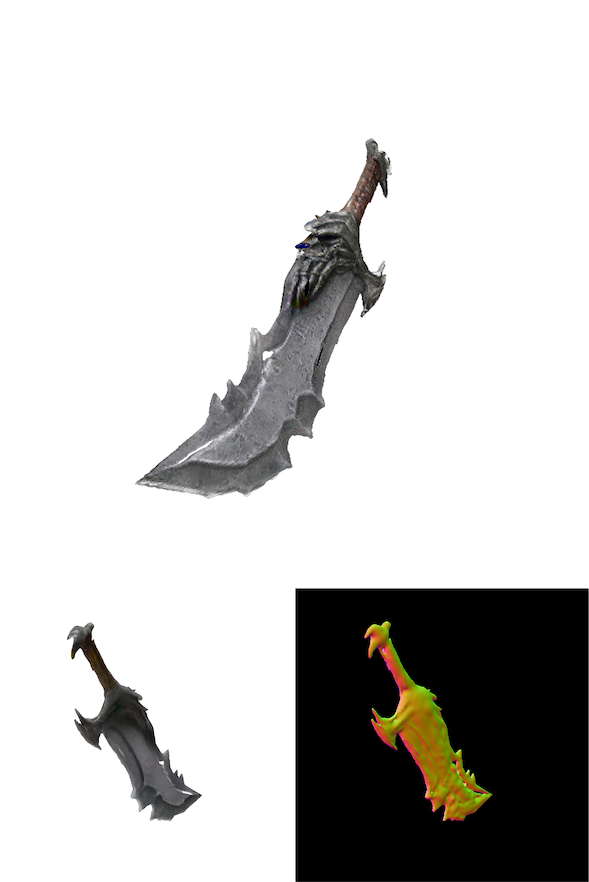} 
        \end{minipage} & 
        \begin{minipage}{.14\textwidth}
          \includegraphics[width=\linewidth]{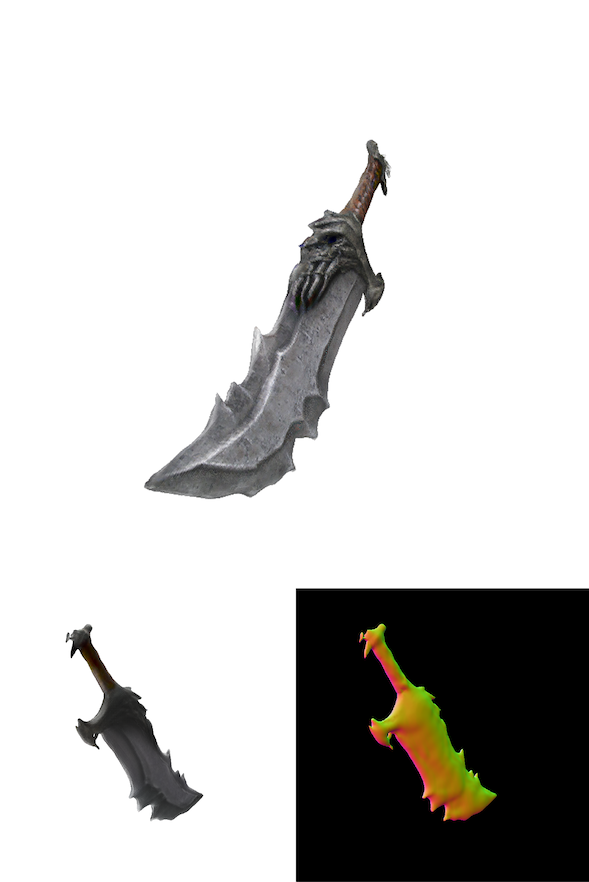}\end{minipage} \\
        &  \cellcolor{Gainsboro!60} 11 min 38 sec & \cellcolor{Gainsboro!60}\textbf{3 min 45 sec} & & \cellcolor{Gainsboro!60} 6 min 1 sec & \cellcolor{Gainsboro!60}\textbf{1 min 13 sec } \\
        \begin{minipage}{.14\textwidth}
          \includegraphics[width=\linewidth]{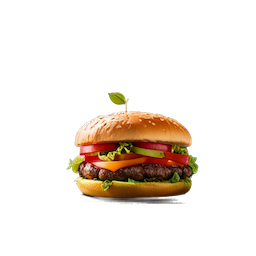} 
        \end{minipage} & 
        \begin{minipage}{.14\textwidth}
          \includegraphics[width=\linewidth]{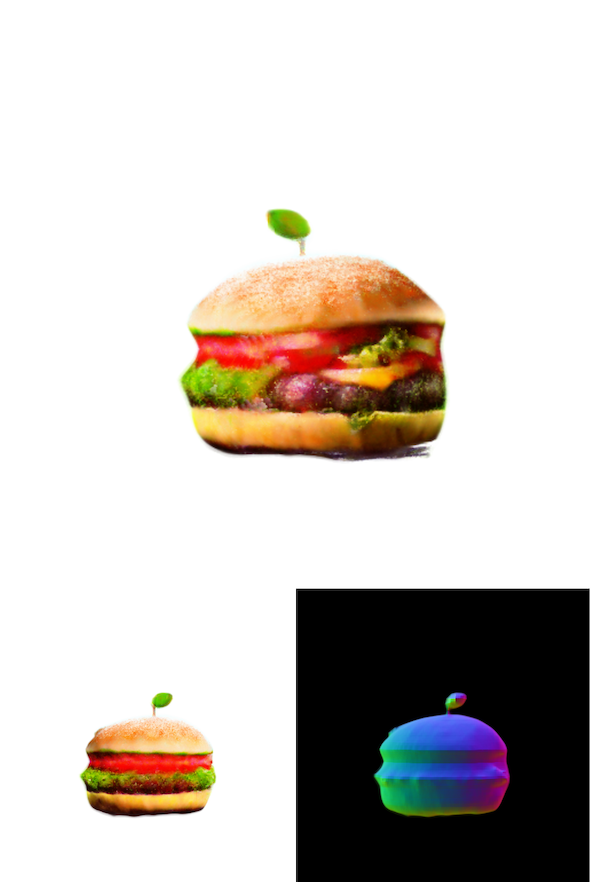} 
        \end{minipage} & 
        \begin{minipage}{.14\textwidth}
          \includegraphics[width=\linewidth]{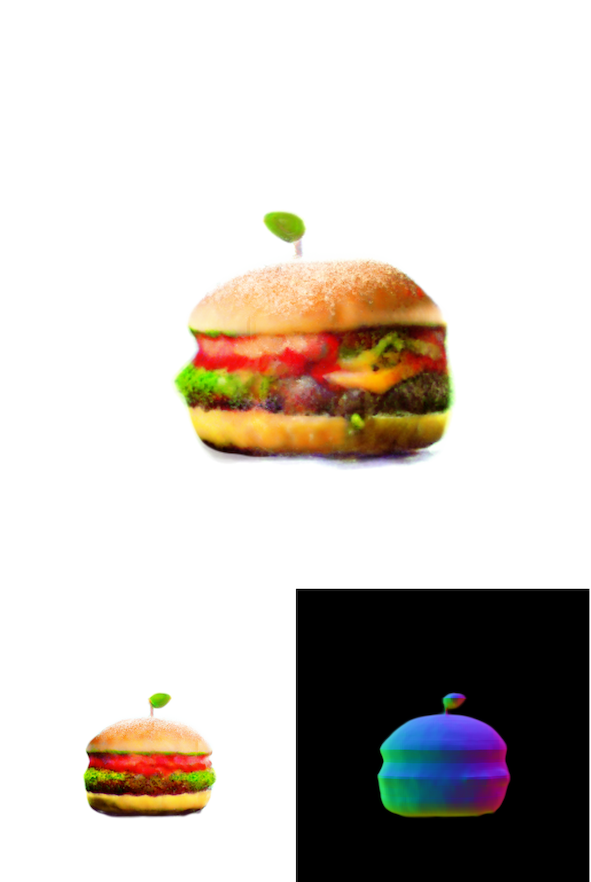}\end{minipage}  &
          \hfill
         \begin{minipage}{.14\textwidth}
          \includegraphics[width=\linewidth]{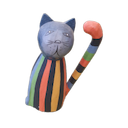} 
        \end{minipage} & 
        \begin{minipage}{.14\textwidth}
          \includegraphics[width=\linewidth]{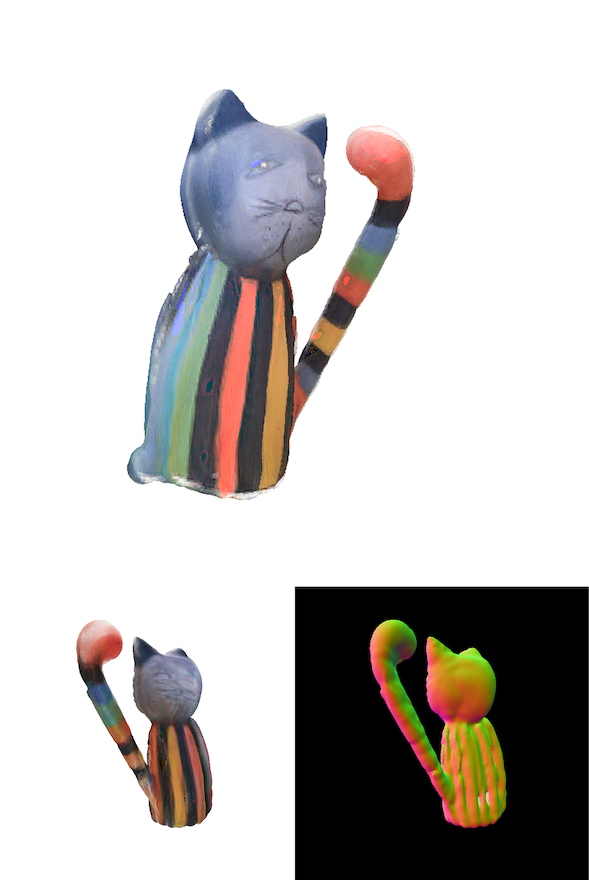} 
        \end{minipage} & 
        \begin{minipage}{.14\textwidth}
          \includegraphics[width=\linewidth]{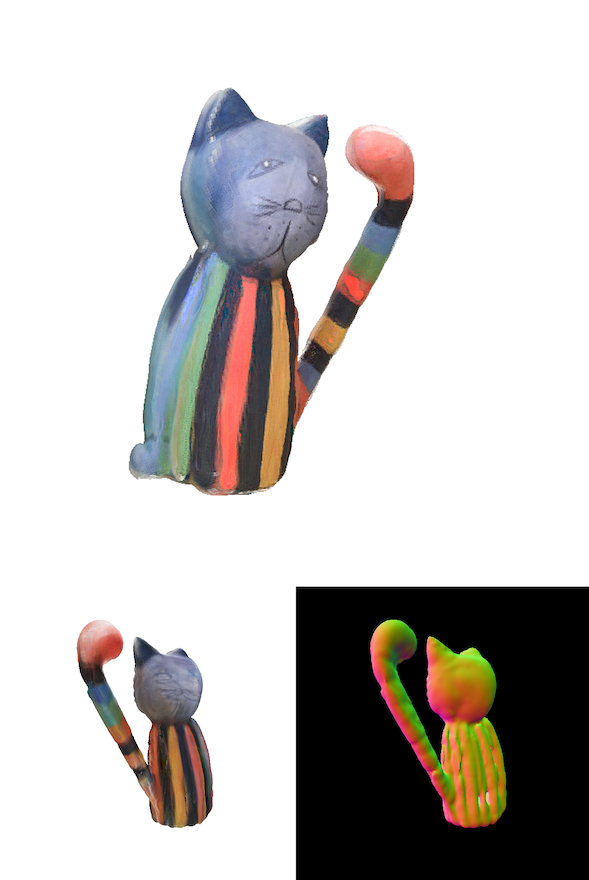}\end{minipage} \\
        &  \cellcolor{Gainsboro!60} 10 min 53 sec & \cellcolor{Gainsboro!60}\textbf{3 min 17 sec} & & \cellcolor{Gainsboro!60} 5 min 56 sec & \cellcolor{Gainsboro!60}\textbf{1 min 15 sec} \\
        \begin{minipage}{.14\textwidth}
          \includegraphics[width=\linewidth]{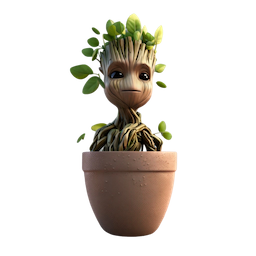} 
        \end{minipage} & 
        \begin{minipage}{.14\textwidth}
          \includegraphics[width=\linewidth]{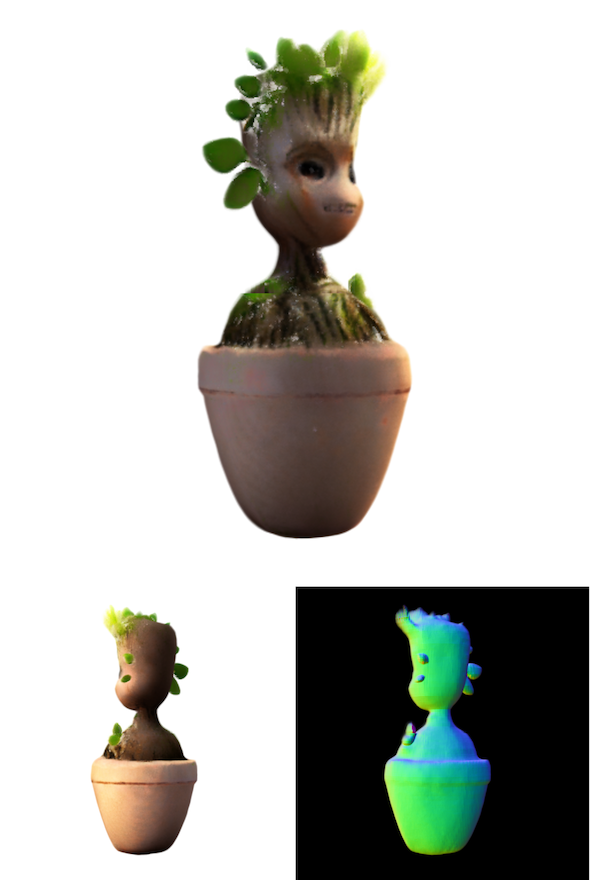} 
        \end{minipage} & 
        \begin{minipage}{.14\textwidth}
          \includegraphics[width=\linewidth]{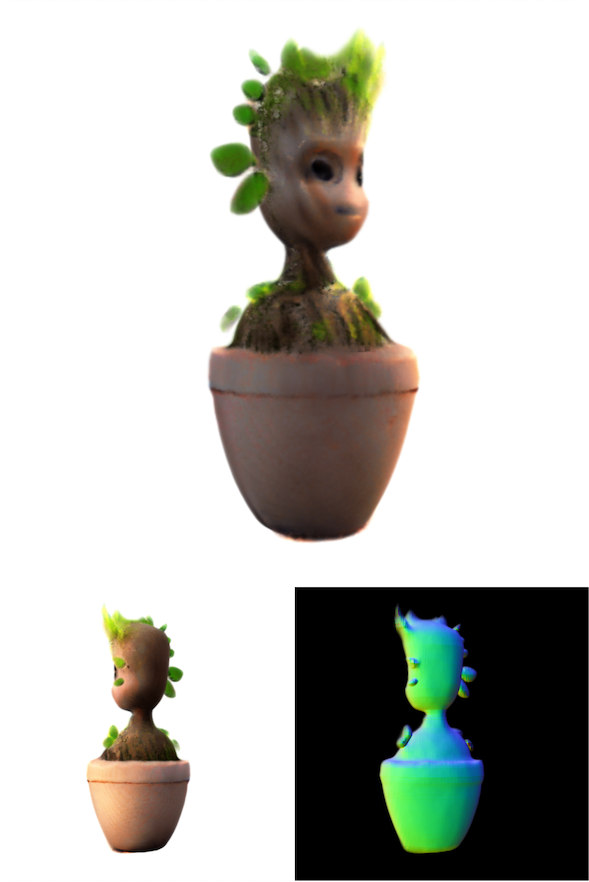}\end{minipage}  &
          \hfill
         \begin{minipage}{.14\textwidth}
          \includegraphics[width=\linewidth]{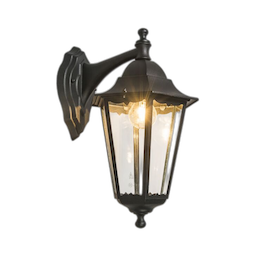} 
        \end{minipage} & 
        \begin{minipage}{.14\textwidth}
          \includegraphics[width=\linewidth]{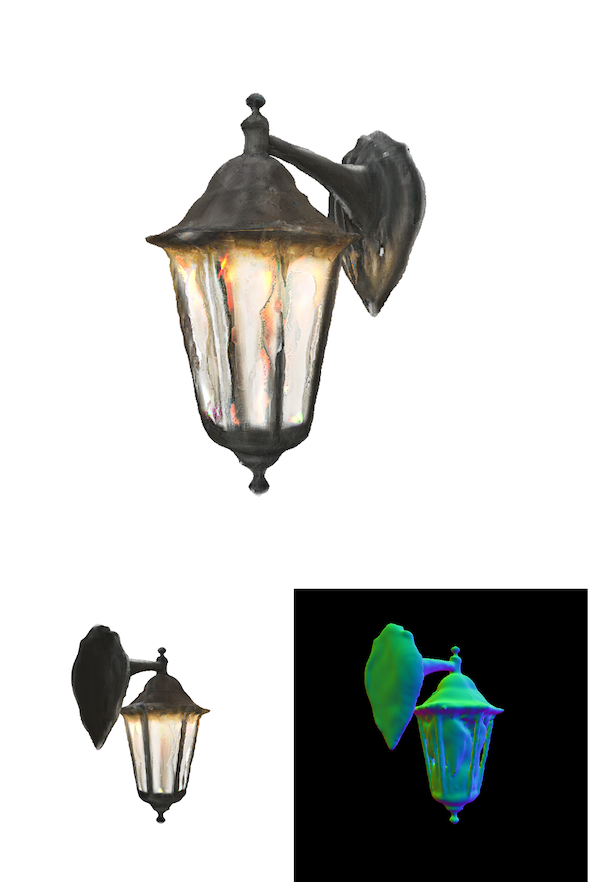} 
        \end{minipage} & 
        \begin{minipage}{.14\textwidth}
          \includegraphics[width=\linewidth]{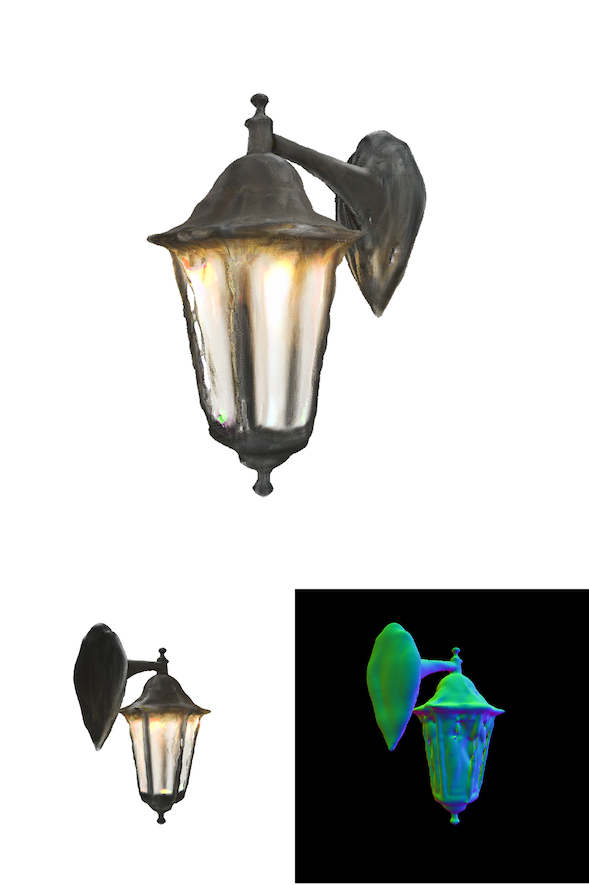}\end{minipage} \\
        &  \cellcolor{Gainsboro!60} 10 min 51 sec & \cellcolor{Gainsboro!60}\textbf{3 min 11 sec} & & \cellcolor{Gainsboro!60} 5 min 58 sec & \cellcolor{Gainsboro!60}\textbf{1 min 15 sec}\\
          \rowcolor{Gainsboro!60}
      Avg. Speedup  &  \multicolumn{2}{c}{3.28x} &  Avg. Speedup & \multicolumn{2}{c}{4.8x}   \\
         
     \end{tabular}
    \caption{Visual comparisons for Image-to-3D generation using Zero-1-to-3~\citep{Liu2023-kk}. Our method achieves more than 3x and 4x average speedup respectively when applied to Zero-1-to-3 while achieving almost identical generation results.}
    \label{fig:image-to-3d}
\end{figure*}

\section{Accelerating Image-to-3D Generation}\label{app:image-to-3d}

Many works have also explored score distillation for Image-to-3D generation using 2D-diffusion finetuned on view-dependent data~\citep{Liu2023-kk,Qian2023-rv,Liu2023-ut}. Among the most popular approaches is Zero-1-to-3~\citep{Liu2023-kk}, which finetunes a large-scale diffusion model for novel-view synthesis given a single image and novel-view embeddings. It also serves as a powerful 3D-aware prior for score distillation, which luckily our framework can be directly applied to. In this section, we investigate our framework's application to the Image-to-3D generation task.

For evaluation, we choose NeRF~\citep{Mildenhall2021-wv} and 3D Gaussian Splatting~\citep{Kerbl2023-kn} as the two representative examples for 3D representations with constant and changing dimensions during optimization. Each representation is equipped with Zero-1-to-3 and a source image for generating a novel 3D object, and we show that our framework can achieve substantial speedup when applied to Zero-1-to-3 while retaining generation quality. For both representations, we use batch size 16 unless otherwise noted, and we run SDS for 1200 steps and 500 steps respectively (details in \appref{app:exp}). We show 3 examples for each representation in~\figref{fig:image-to-3d}, where we compare novel views of the 3D results from Zero-1-to-3 and the results from Zero-1-to-3 accelerated by \model. 

We observe that our framework can achieve almost identical generation output with much shorter runtime for both representations. The wallclock time speedup is consistently more than 3x and 4x the original runtime. NeRF achieves lower speedup than the Text-to-3D counterpart due to the lower number of total steps compared to Text-to-3D generation (\eg 25,000 steps), so the time for the costly initial model and data preloading for all GPUs is not effectively amortized. 3D Gaussian Splatting is less affected thanks to its lightweight representation conducive to fast cross-GPU communication and  DreamGaussian's~\citep{Tang_undated-od} simple implementation with minimal initial allocation cost. 

\fi

\end{document}